\documentclass[review]{elsarticle}
\makeatletter
\def\ps@pprintTitle{%
 \let\@oddhead\@empty
 \let\@evenhead\@empty
 \def\@oddfoot{}%
 \let\@evenfoot\@oddfoot}
\makeatother
\usepackage{lineno,hyperref}
\usepackage{color,soul}
\usepackage{todonotes}
\usepackage{booktabs}
\usepackage{multirow}
\usepackage{longtable}
\usepackage{changebar}
\usepackage{graphicx}
\usepackage[caption=false]{subfig}

\modulolinenumbers[5]
\journal{}

\begin{document}

\begin{frontmatter}

\title{Interpretable Machine Learning Classifiers for Brain Tumour Survival Prediction}

\author[1]{Colleen E. Charlton\footnote[1]{Current address for C.E. Charlton: Krembil Centre for Neuroinformatics, Centre for Addiction and Mental Health, Toronto, Ontario, Canada}}
\ead{Colleen.Charlton@camh.ca}

\author[2,3,4,5]{Michael Tin Chung Poon}
\ead{michael.poon@ed.ac.uk}

\author[2,3,4]{Paul M. Brennan}
\ead{paul.brennan@ed.ac.uk}

\author[1]{Jacques D. Fleuriot}
\ead{jdf@ed.ac.uk}


\address[1]{Artificial Intelligence and its Applications Institute, School of Informatics, University of Edinburgh, 10 Crichton Street, Edinburgh, EH8 9AB, UK}
\address[2]{Cancer Research UK Brain Tumour Centre of Excellence, CRUK Edinburgh Centre, University of Edinburgh, Edinburgh, UK}
\address[3]{Department of Clinical Neuroscience, Royal Infirmary of Edinburgh, 51 Little France Crescent, EH16 4SA, UK.}
\address[4]{Translational Neurosurgery, Centre for Clinical Brain Sciences, University of Edinburgh, Edinburgh, UK}
\address[5]{Centre for Medical Informatics, Usher Institute, University of Edinburgh, Edinburgh, UK}

\begin{abstract}
Prediction of survival in patients diagnosed with a brain tumour is challenging because of heterogeneous tumour behaviours and responses to treatment. Better estimations of prognosis would support treatment planning and patient support. Advances in machine learning have informed development of clinical predictive models, but their integration into clinical practice is almost non-existent. One reasons for this is the lack of interpretability of models. In this paper, we use a novel brain tumour dataset to compare two interpretable rule list models against popular machine learning approaches for brain tumour survival prediction. All models are quantitatively evaluated using standard performance metrics. The rule lists are also qualitatively assessed for their interpretability and clinical utility. The interpretability of the “black box" machine learning models is evaluated using two post-hoc explanation techniques, LIME and SHAP. Our results show that the rule lists were only slightly outperformed by the black box models. We demonstrate that rule list algorithms produced simple decision lists that align with clinical expertise. By comparison, post-hoc interpretability methods applied to “black box" models may produce unreliable explanations of local model predictions. Model interpretability is essential for understanding differences in predictive performance and for integration into clinical practice. 

\end{abstract}

\begin{keyword}
Bayesian rule lists, interpretable models, machine learning, brain cancer, survival
\end{keyword}

\end{frontmatter}


\section{Introduction}

Glioblastomas (GBM) are the most common malignant brain tumour and have the poorest outcomes. Average survival is 12-18 months and the 5-year survival rate is less than 5\% \cite{poon2020longer}. Lower grade gliomas have an average survival of 7 years, but ultimately most progress to GBM and death \cite{claus2015survival}. An accurate prediction of prognosis for patients would inform treatment planning and patient support, but this is challenging. Various factors impact prognosis, but the precise contribution of each factor, and combination of factors to outcomes appears to vary between patients.

 At the group level, basic statistical models are well-established in brain tumour survival analysis, but patient level survival prediction remains a challenge, possibly because of the well described biological heterogeneity  of the disease and of treatment responses. Several survival studies have used the Cox proportional hazards (Cox PH) model \cite{cox1984analysis}, a popular survival model in statistics, to identify relevant clinical features for the construction of a brain tumour nomogram \cite{barnholtz2012nomogram, gittleman2017independently, gittleman2020independently, gorlia2008nomograms} which graphically depict a statistical model to be used to estimate individualised cancer prognosis \cite{iasonos2008build}. However, simple multivariable regression techniques are ineffective at identifying novel informative patterns from data \cite{bzdok2018points}, so machine learning (ML) approaches are increasingly being explored in brain tumour survival analyses \cite{bakas2018identifying, jain2014outcome} (see also Kourou et al.\ \cite{kourou2015machine} and Wang et al.\ \cite{wang2019machine} for general reviews). Such studies often use genetic or imaging data, with complex black box models to make prognostic predictions. In clinical practice this type of data is seldom, hence these models are of little use. Fulop et al.\ \cite{Fulop2019predicting} used a large clinical and molecular brain tumour dataset to predict 400-, 900- and greater than 900-day survival after surgery using several ML methods. The authors found that a neural network model performed the best with an accuracy of 59\%. Furthermore, the authors used LIME \cite{Ribeiro2016} to understand the main drivers behind the wrong predictions and emphasized the importance of model interpretability for clinical decision-making. Senders et al.\ \cite{senders2020online} recently used demographic, socioeconomic, clinical, and radiographic features for the creation of an online calculator for the prediction of glioblastoma survival. The authors compared 15 ML and statistical algorithms and an Accelerated Failure Time \cite{wei92aft} algorithm was selected. However, a follow-up Letter to the Editor in the same journal \cite{urso2020letter} noted inconsistencies in the model calculations and highlighted the danger of non-healthcare professionals accessing this online resource which may provide misleading information. Although ML can be used to build predictive models with superior performance and generalisability \cite{wang2019machine, song2004comparison}, the implementation of such models in a clinical setting can come with safety, legal and ethical considerations. In healthcare, there is a desire for interpretable and explainable AI/ML models to give end users (e.g.\ clinicians) the support that will allow them to accept or reject predictions, thus enabling them to make informed judgements when it comes to high-stake medical decisions (see Ahmad et al. \cite{ahmad2018interpretable} and Holzinger et al. \cite{holzinger2017we} for reviews on interpretable ML in healthcare). 

There is no all-purpose definition of interpretability since this is a subjective concept that is often domain-specific \cite{Rudin2019, ruping2006learning}. One way to define interpretability is as the degree to which a human can understand the cause of a decision \cite{Miller2019}. Thus a model $M_1$ may be considered more interpretable than a model $M_2$ if $M_1$'s decisions are easier to comprehend than those of $M_2$ \cite{carvalho2019machine}. Interpretability may be achieved by either using an intrinsically interpretable model whereby its simple structure allows end-users to understand feature relationships and final predictions, or by applying post-hoc explanation techniques to analyse and extract information from a trained model \cite{Molnar2019}. Most ML models are not originally designed to be interpretable and advancement in ML performance has led to the belief in a model's accuracy-interpretability trade-off \cite{Rudin2019}. However, interpretability may be used as a tool to improve accuracy \cite{carvalho2019machine} and models with interpretability constraints have already been shown to perform on par with unconstrained models across several healthcare domains \cite{Caruana2015, Razavian2015, Rudin2018}. 

In this paper we explore the use of Bayesian Rule Lists (BRL) \cite{Letham2015} and Falling Rule Lists (FRL) \cite{Wang2015} as two types of intrinsically interpretable ML models and apply them to the prediction of patient survival using a novel brain tumour dataset. Both models combine pre-mined frequent patterns from the dataset into a decision list using Bayesian statistics \cite{Letham2015}. The FRL model is an extension of the BRL algorithm which produces an ordered decision list whereby the estimated probability of success decreases down the list \cite{Wang2015}. The BRL and FRL algorithms are compared to a baseline Cox PH model and popular black box models by looking at the prediction for brain tumour survival. To represent the class of black box methods, we chose random forest (RF) \cite{breiman2001random}, logistic regression (LR) \cite{menard2002applied} and support vector machine (SVM) \cite{suykens1999least}, each of which have varying degrees of interpretability. Finally, in an attempt to understand the decisions made by non-transparent models, post-hoc interpretability methods LIME (Local Interpretable Model-Agnostic Explanations) \cite{Ribeiro2016} and SHAP (SHapley Additive exPlanations) \cite{Lundberg2017} are applied to the black box models. For a given instance (e.g.\ patient), the explanations produced by the post-hoc methods are compared between the three ML models.

The remainder of this paper is organised as follows: In Section \ref{dataset} we introduce both the raw and final dataset and the required preprocessing techniques; Section \ref{methods} outlines model construction; in Section \ref{results} we report the model's quantitative and qualitative results; in Section \ref{discussion} we analyse and discuss the results and Section \ref{conclusion} closes with final conclusions.

\section{The Brain Tumour Data} \label{dataset}

\subsection{The Raw Dataset}

This paper explores an anonymised hospital-based brain tumour dataset collected from routine electronic healthcare data of patients who presented to regional neuro-oncology teams in the UK with a brain tumour diagnosis. Initially, the dataset contained 1296 patient records and 225 predictor variables. A preliminary exploratory analysis of the data led to the removal of incomplete variables (features less than 60\% filled) ($n$=179) and variables irrelevant to predicting outcomes (e.g. location of first imaging, clinician ordering CT (open access CT), contrast agent) ($n$=14), as well as the grouping of duplicated predictor variables (e.g. Symptom 1 and Symptom 1 - other) ($n$=10). Additionally, a number of patients were removed due to incomplete records (i.e. a patient is missing more than 60\% of the reduced predictor variables) ($n$=51) or because of a lack of symptomatology information (i.e. a patient did not present with any symptoms or signs) ($n$=227). The remaining 1018 patients records contained 21 predictor variables and one dependent variable, namely patient survival in days.
 
The raw dataset contained significant heterogeneity with more than 30 different brain tumour types, the most common being glioblastoma ($n$=540), followed by metastasis ($n$=198), glioma ($n$=186) and meningioma ($n$=174). The minimum age of diagnosis is 16 years while the oldest is 97 years, and a median age of 61 years. There are an almost equal number of male and female patients (51\% and 49\%, respectively). 18\% of patients had a previous history of cancer and 48\% of patients presented with a co-morbidity, the most common being cardiovascular (15\% of all patients). The most common location for a tumour was in the frontal lobe (34\%) followed by the temporal lobe (22\%). More than half of the patients (68\%) had some type of surgery and 23\% of patients underwent chemotherapy. This is inline with current treatment protocols where surgery is often the first line of treatment, followed by chemotherapy and concurrent radiotherapy for malignant tumours \cite{Stupp2005}. Finally, 23\% of patients received no treatment, reflecting either the benign nature of the tumour, or conversely its advanced state or poor clinical condition of the patient. 

Patient survival was measured in days from radiological diagnosis of brain tumour and 35\% of patients were still alive at the time of dataset analysis. This is known as \textit{censored data}, which is common in survival analysis, whereby the value of an observation, in this case survival, is only partially known \cite{klein2006survival}. In the raw dataset, the largest survival time is 3964 days, or about 10 years, however only 11 patients have a survival time greater than 2000 days. At the time of data analysis, all patients with censored survival data had been alive for more than one year following diagnosis.

\subsection{Preprocessing}

\begin{figure}[h]
    \centering
    \includegraphics[width=0.8\linewidth]{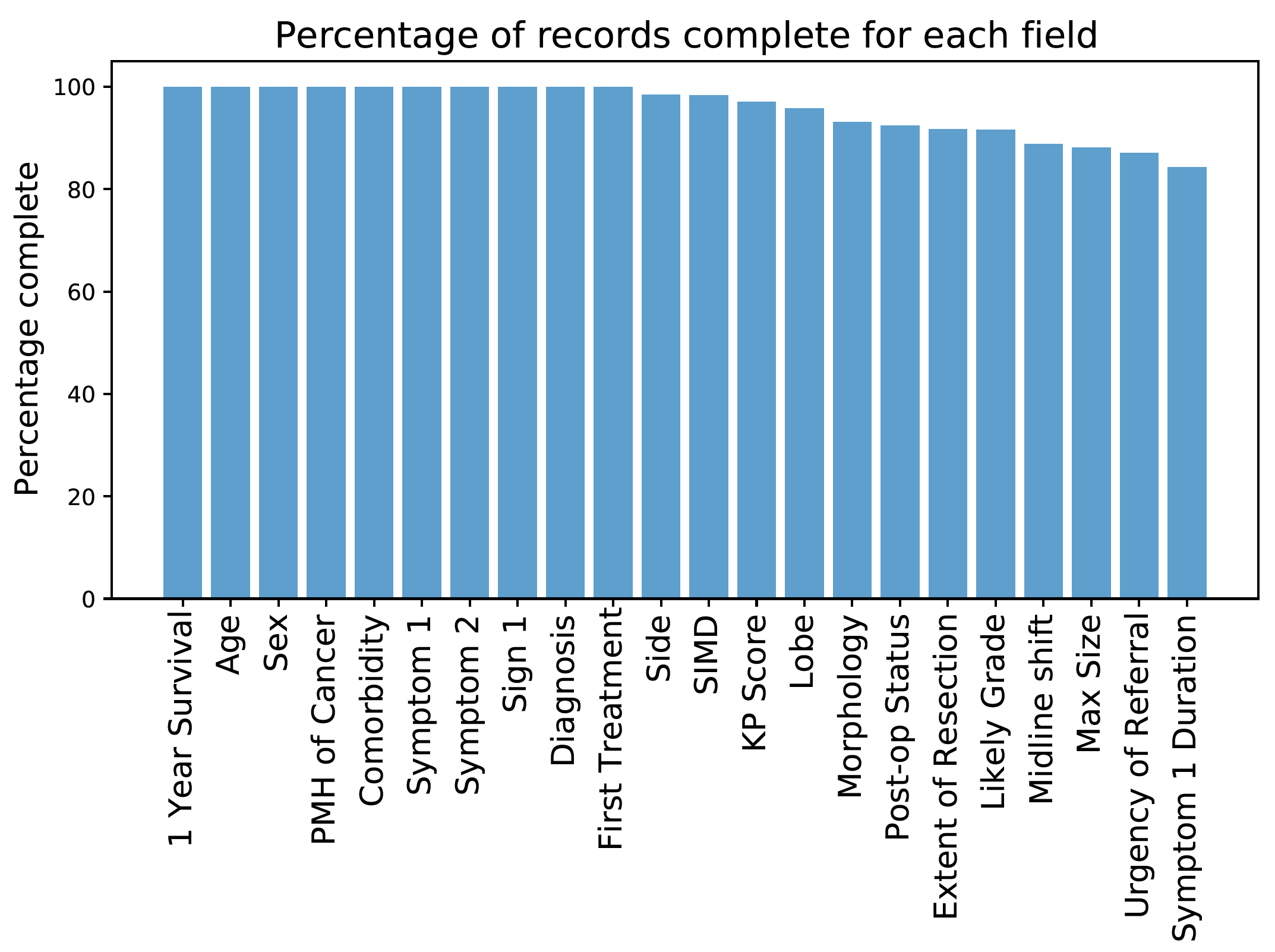}
    \caption{Percentage of records complete for each feature in the raw dataset. See the Appendix for feature descriptions.}
    \label{records_complete}
\end{figure}

Next, we review various pre-processing steps that were needed to get the narrowed-down dataset of 1018 patients into a state suitable for our analyses. Figure \ref{records_complete} illustrates the percentage of patient records complete for each feature.

\subsubsection{Imputation}

Given the small dataset size, missing data was managed though imputation rather than deletion. Treating all missing data the same would be a strong oversimplification as missing data can come from a variety of sources \cite{sterne2009multiple}. An entry for a feature may be absent, but this does not imply that the entry is truly missing. For example, a patient may only present with one symptom thus leaving the remaining symptom features empty. This creates the appearance of missing data but in fact the empty entries are correctly missing. The variables to be imputed were all believed to be missing completely at random.

A number of imputation techniques were tried to determine which method was appropriate for each incomplete variable. In particular, a baseline mode- and mean-fill was used for categorical and continuous variables, respectively, and were compared to a $k$-nearest neighbours ($k$-NN) \cite{beretta2016nearest} and regression \cite{allison2000multiple} imputation technique. For $k$-NN, the use of normalised and non-normalised features were both explored to account for differences in numerical values. The continuous variables in the dataset are constrained thus outliers were not a concern (e.g. the feature \emph{Symptom 1 Duration} must fall between 0 and 52 weeks). For each variable with missing values, the imputation techniques were evaluated on the entire dataset using 10-fold cross-validation. The imputation of categorical and continuous variables was assessed using accuracy and the standard mean square error, respectively. The optimal imputation method for each variable was then implemented on the full dataset, whereby the missing variables were replaced with the model's predicted output. More information about the imputation methods are described in a forthcoming paper based on the following thesis \cite{May2019}.

\subsubsection{Discretisation}

Following imputation, all continuous variables (\textit{Age}, \textit{Symptom 1 Duration}, and \textit{Maximum Tumour Size}) were discretised in order to support the association rule mining \cite{zhang2003association} employed by the BRL and FRL algorithms (see Section \ref{rule_lists}). Most rule-mining approaches make the (restrictive) assumption that all features are binary or categorical. As part of the discretisation process, we compared three methods: uniform binning, quantiles and $k$-means. Each feature was divided into groups ranging from size 2 to 12, increasing by increments of two. We chose the maximum to be 12 as it was a natural divider for the feature \textit{Maximum Tumour Size}, which contained the largest range in values (0 - 120). The accuracy of the discretised features was compared to the continuous feature, using the BRL model with default hyperparameters. Classification of patient survival greater than one year was evaluated on the entire dataset using 5-fold-cross validation. 

\subsubsection{Dealing with Feature Collinearity}
Finally, in our dataset, two sets of features were found to be highly correlated: \textit{Tumour Type} and \textit{Likely Grade} as well as \textit{First Treatment} and \textit{Extent of Resection}. Tumour type refers to the kind of tumour a patient is diagnosed with, while likely grade is an indicator of how quickly a tumour is likely to grow or spread. A grade I/II tumour is benign (slow growing and unlikely to spread within the brain) and a grade III/IV tumour is malignant (fast growing and likely to spread within the brain). However, by definition, a glioblastoma is a grade IV glioma tumour. Thus to reduce collinearity in the data, Tumour Type and Likely Grade were combined into a single feature called \textit{Diagnosis}. Tumour types were separated into benign (or low-grade) and malignant (or high-grade) categories (e.g.\ Glioma Benign and Glioma Malignant). Metastatic tumours encompass a mix of tumour types that originate elsewhere in the body. By the fact that they migrate to the brain, metastatic tumours are all aggressive and thus classified as malignant. The second set of features, \textit{First Treatment} and \textit{Extent of Resection} (EOR), also suffer from multicollinearity. EOR refers to the amount of cancerous cells removed during surgery (e.g.\ 90-99\%) and is only relevant if the first treatment a patient receives is surgery, which is not always the case. Thus we integrated EOR information into First Treatment (e.g.\ Surgery 100\%, Surgery 90-99\%) to create a more informative feature - the feature name remained \textit{First Treatment}. 

\subsubsection{The Final Dataset} \label{sec:final_data}

Our final dataset contains 1018 patient records and 19 predictor variables including patient demographics (e.g.\ sex, age), medical history (e.g.\ history of cancer, comorbidity), symptom features (e.g.\ symptom types and duration), radiological tumour analysis (e.g.\ diagnosis, morphology) and treatment details (e.g.\ first treatment, post-op performance status). Many of these prognostic factors have been well-documented in the literature including age, Karnofsky performance (KP) score, symptoms, morphology, diagnosis (i.e.\ tumour type, likely grade) and treatment \cite{gehan1977prognostic, mckinney2004brain, walid2008prognostic, gittleman2018survivorship, lapointe2018primary}. Table \ref{data_variables_salient} provides a detailed summary of the salient variables in the final dataset, including their descriptions, value and percentage of each value present in the dataset following imputation and discretisation (see Table \ref{tab:long} in the Appendix for a comprehensive overview of all predictor variables in the final dataset). We briefly discuss some of the salient features below and some final processing of the features.

{\footnotesize
\begin{longtable}{p{2cm}p{4.2cm}p{3.3cm}p{1.3cm}}

\caption{Overview of 9 salient dataset variables including their descriptions, value and percentage of each value present in the final dataset following imputation and discretisation.} \label{data_variables_salient} \\

\toprule
Name & Description & Value & Proportion (\%) \\* \midrule
\endfirsthead

\multicolumn{4}{c}%
{{\bfseries \tablename\ \thetable{} -- continued from previous page}} \\
\toprule
Name & Description & Value & Proportion (\%) \\* \midrule
\endhead

\multicolumn{4}{r}{{Continued on next page}}\\ \bottomrule
\endfoot
\bottomrule
\endlastfoot

\multirow{4}{*}{Age} & \multirow{6}{*}{the age of a patient} & 0-44 & 17.1 \\
 &  & 45-54 & 18.7 \\
 &  & 55-61 & 16.0 \\
 &  & 62-67 & 15.8 \\
 &  & 68-74 & 16.9 \\
 &  & 75$+$ & 15.5 \\* \midrule
 
\multirow{2}{*}{Sex} & \multirow{2}{*}{the sex of the patient} & Male & 50.5 \\
 &  & Female & 49.5 \\* \midrule

\multirow{4}{*}{\begin{tabular}[c]{@{}l@{}}Karnofsky \\ Performance \\ Score \\ (KP Score)\end{tabular}} & \multirow{4}{*}{\begin{tabular}[c]{@{}l@{}}a common measure in oncology \\ to assess the functional state \\ of a patient (see Figure \ref{kps_table} in \\ Appendix) \end{tabular}} & 100 & 37.6 \\
 &  & 90 & 28.6 \\
 &  & 80 & 14.6 \\
 &  & $\leq$70 & 19.2 \\* \bottomrule
 
 \pagebreak
 
\multirow{6}{*}{Symptom 1} & \multirow{6}{*}{\begin{tabular}[c]{@{}l@{}}the first symptom type a \\  patient presented with \\ (reported by the patient)\end{tabular}} & Focal Neurology & 34.6 \\
 &  & Headache & 28.4 \\
 &  & Fits/Faints/Falls & 17.1 \\
 &  & Behavioural/Cognitive & 16.7 \\
 &  & Other/Non-specific & 2.4 \\
 &  & Non-specific Neurological & 0.8 \\* \midrule
 
 \multirow{6}{*}{Sign 1} & \multirow{6}{*}{\begin{tabular}[c]{@{}l@{}}the first sign type a patient \\ presented with \\ (observed by the physician)\end{tabular}} & No Signs & 42.7 \\
 &  & Neurological & 36.2 \\
 &  & Cognitive & 15.0 \\
 &  & Cranial Nerve & 5.0 \\
 &  & Other & 0.8 \\
 &  & Behavioural & 0.3 \\* \midrule
 
\multirow{9}{*}{\begin{tabular}[c]{@{}l@{}}Diagnosis \\ (or Tumour \\ Type)\end{tabular}} & \multirow{9}{*}{\begin{tabular}[c]{@{}l@{}}the type of brain tumour a \\ patient was diagnosed with\end{tabular}} & Glioma Malignant & 46.5 \\
 &  & Metastasis & 19.0 \\
 &  & Meningioma Benign & 13.6 \\
 &  & Glioma Benign & 7.1 \\
 &  & Rare Tumour Benign & 4.7 \\
 &  & Lymphoma Malignant & 4.1 \\
 &  & Meningioma Malignant & 2.3 \\
 &  & Rare Tumour Malignant & 1.5 \\
 &  & Hemangioblastoma Benign & 1.2 \\* \midrule
 
 \multirow{3}{*}{Morphology} & \multirow{3}{*}{\begin{tabular}[c]{@{}l@{}}the histological classification \\ of the tumour based on the \\ cell types present\end{tabular}} & Heterogenous & 68.5 \\
 &  & Homogenous & 31.5 \\
  &  &  & \\* \midrule
 
 \multirow{7}{*}{\begin{tabular}[c]{@{}l@{}}Post-operative \\ Performance \\ Status\end{tabular}} & \multirow{7}{*}{\begin{tabular}[c]{@{}l@{}}a measure of a patient’s level \\ of functioning following surgery \\ in terms of their ability for \\ self-care, daily activity, and \\ physical ability (see Table \ref{postop_score} \\ in Appendix)\end{tabular}} & 0 & 31.5 \\
 &  & 1 & 27.4 \\
 &  & 2 & 6.2 \\
 &  & 3 & 1.9 \\
 &  & 4 & 1.4 \\
 &  & 5 & 0.2 \\
 &  & No Surgery & 31.4 \\* \midrule
 
 \multirow{9}{*}{\begin{tabular}[c]{@{}l@{}} First Treatment \end{tabular}} & \multirow{9}{*}{\begin{tabular}[c]{@{}l@{}}the type of first cancer \\ treatment \end{tabular}} & Surgery Removal 100\% & 16.0 \\
 &  & Surgery Removal 90-99\% & 24.4 \\
 &  & Surgery Removal 50-89\% & 6.4 \\
 &  & Surgery Removal $<$50\% & 4.9 \\
 &  & Biopsy & 16.9 \\
 &  & Radiotherpay & 5.5 \\
 &  & Chemotheapy & 0.9 \\
 &  & Other (e.g.\ steroids) & 2.5 \\
 &  & No Treatment & 22.5 \\* \bottomrule
 
\end{longtable}}

\textbf{KP Score:} This is a standard way of assessing a patient's ability to perform everyday tasks \cite{karnofsky1948use}. The scale is a `gold standard' in clinical oncology and is commonly used to determine a cancer patient's expected tolerance to treatments (e.g.\ chemotherapy). The scores ranges from 0 (dead) to 100 (normal) and is scored in deciles, although the values are ordinal (see Table \ref{kps_table} in the Appendix for the original definition of the KP scores). This means that a value assigned to a patient is based on a ranking but the numerical value associated with this rank is not meaningful. Thus the difference between the values 70 and 90 is not equivalent to the difference between the values 40 and 60. Furthermore, the KP scale may be subject to bias \cite{frappaz2020assessment}. A patient's KP score is determined by clinicians, and when compiling a dataset this can result in inter-observer subjectivity \cite{Taylor1999, Sorensen1993}. To reduce the bias associated with the KP score, values of 70 and below were aggregated due to their negative association with survival \cite{chaichana2010proposed} (a KP score of 70 reflects someone who can `care for self, but who is unable to carry on normal activity or to do active work'). KP scores of 80 and above remained separate allowing for a more fine-grained analysis of the values associated with survival. 

\textbf{Symptom 1:} A symptom is observed by the patient themselves (subjective) and is often what drives a patient to consult a physician. Symptom 1 refers to the first symptom a patient presents with. The symptom data in the raw dataset had a high cardinality of 37 different symptom types, with many of these types pertaining to a small number of patients. Thus we decided to group symptom types into six overarching categories -- e.g.\ Headache, Fits/Faints/Falls and Behavioural/Cognitive -- based on work by Ozama et al.~\cite{Ozawa2019} to create a more homogeneous set of symptom types. An outline of the symptom groupings are summarised in Table \ref{symp_domain} of the Appendix. 

\textbf{Sign 1:} A \emph{sign} is observed by a physician (objective). Sign 1 refers to the first sign a patient presents with. The sign data in the raw dataset also had a high cardinality (26 different types), thus the data was additionally grouped into six larger domains -- e.g. neurological and cognitive--  based on the advice of the consulting clinical experts (see Table \ref{sign_grouping} in the Appendix). Although all patients in the final reduced dataset presented with at least one symptom, 43\% of patients did not present with any signs. 

\textbf{Diagnosis (Tumour Type):} Brain tumours are broadly named based on the type of normal cell that they most resemble, and their location in the brain \cite{webtumourtypes}. In the raw dataset the tumour types had a high cardinality with many entries referring to the same general tumour type (e.g. meningioma suprasellar and meningioma at cerebellopontine (CP) angle). Tumour types that appeared in less than 10 patients were grouped into a “Rare Tumour” category. Additionally, the tumour type may be benign (i.e. grade I/II), or malignant (i.e. grade III/IV). In the final dataset, the brain tumour types were reorganised based on type and malignancy (e.g. Glioma Benign, Glioma Malignant), and reduced to a cardinality of 9.

\textbf{Morphology:} Each tumour type can have a different sub-classification, which is represented as morphology. A tumour with heterogenous morphology contains diverse cell types with distinct molecular structure that may have different levels of sensitivity to treatment \cite{dagogo2018tumour}. In comparison, a homogenous tumor contains the same or similar genetic or epigenetic characteristics \cite{marusyk2010tumor}. Hence tumour morphology may be indicative of treatment response.

\textbf{First Treatment:} The first type of cancer treatment a patient receives is based on the presumed type based on imaging, location of the tumour, and the patient's overall health (e.g. KP score $\leq$ 70). Surgery, for example, may be the only treatment necessary depending on the grade of the tumour and extent of resection. Information on extent of resection was included in the first treatment types (e.g. Surgery 100\%, Surgery 90-99\%, Biopsy, etc.). Other treatment types include radiotherapy and chemotherapy.

\section{Methods}\label{methods}

\subsection{Modelling Techniques}

As the black box ML models (RF, LR and SVM) cannot directly handle categorical variables, one-hot encoding was used for all 19 features, which resulted in 94 different feature types. For ease in model feature comparison, one-hot encoded features were also used by the Cox PH model. Due to the small dataset size, nested cross validation \cite{varma2006bias} was implemented for hyperparameter tuning and model assessment. The inner loop, which is responsible for the model selection, used 3-fold cross-validation, and the outer loop, which is responsible for estimating model performance, used 5-fold cross-validation. Nested cross validation was repeated for three different random seeds resulting in a $3\times5\times3$ setup. For each model, the average accuracy \cite{bekkar2013evaluation}, macro-F1 \cite{hossin2015review}, and area under the receiver operating characteristic curve (AUROC) \cite{bradley1997use} were reported. All models performed binary classification and one year survival labels were created from the survival data. This resulted in a relatively even split of the dataset: 443 patients (44\%) survived less than a year and 575 patients (56\%) survived more than a year.

\subsection{Cox PH Model} \label{cox_model}

The Cox PH model is a popular regression model for survival analysis \cite{cox1984analysis}. This semi-parametric model predicts the risk of an event occurring (e.g. death) as a function of time. The hazard function $\lambda (t)$ estimates the effect of observed variables $x_k$ on a baseline hazard function $\lambda_0 (t)$. The model can be expressed as: $\lambda (t) = \lambda_0 (t) \textnormal{exp}(\beta_1 x_1 + \beta_2 x_2 + ... + \beta_k x_k)$, where $\beta_1, ... \beta_k$ are the coefficients estimated from the data of the observed variables $x_1, ... x_k$. However, the Cox PH model makes the restrictive assumption that each predictor variable has a multiplicative effect on the baseline hazard function that remains constant over time. Thus nonlinear relationships between predictor variables cannot be modelled by this approach. 

The Cox PH model was implemented using the Python-package lifelines \cite{lifelines} and served as an alternative statistical baseline for the ML models. Due to the one-hot encoding of the categorical variables, the resulting dataset incurred problems with high collinearity. That is, some of the independent variables were highly correlated which tends to inflate the estimated regression coefficients. Thus the first category per feature was dropped - i.e.\ for a variable with k categories, k-1 dummies remained. A penalizer was also added to the model which reduces the size of the coefficients during regression, thus controlling for high correlation and improving the stability of the estimates \cite{ambler2012evaluation}. A search for the optimal penalty parameter between the set of values \{${10^{-3}, 10^{-2},...,10^{1}}$\} was performed and a value of $10^{-2}$ was selected.

The purpose of the Cox PH model is to predict survival time. Thus the model was trained using the continuous survival data and the predicted survival time was converted to binary one year survival labels that were then used to calculate the model's performance. 

\subsection{Rule Lists} \label{rule_lists}

Rule lists are a type of intrinsically interpretable model that produces a series of \textit{if-then} rules, also known as decision (or production) rules, which are used to generate predictions. Each rule is composed of two different sets of items, $a$ and $b$, also known as itemsets. A decision rule is an implication of the form $a \to b$ (or \textit{if...then...}), where $a$ is an antecedent that is followed by a consequent $b$. As an example, for the rule: 
\begin{quote}
{\sf IF Diagnosis: Glioma Malignant AND First Treatment: Surgery 100\% THEN probability of Survival $>$ 1 Year: 90\%}, 
\end{quote}

\noindent the antecedent $a$ is “Diagnosis: Glioma Malignant AND First Treatment: Surgery 100\%”, and the consequent $b$ is “Survival $>$ 1 Year”. If a rule (or set of rules) is satisfied, the model outputs a certain classification. Although the concept of decision rules is well known in AI, more recently, rule-based models have been constructed directly from data with the help of ML \cite{liu2015rule}. 

Instead of being crafted using domain knowledge, rules can be \textit{learned} using frequent itemset mining \cite{agrawal94arm,borgelt2012frequent}, which looks for common patterns in the data. Itemsets, consisting of these frequent patterns, are used to construct the association rules that are normally subject to constraints on minimum support and confidence \cite{agrawal94arm}, although other measures such as lift \cite{hussein2015using} are possible. In particular,
$$
 \textnormal{Support}=\frac{freq(a,b)}{N}
$$
refers to how frequently the itemsets appear in the data, where $freq(a,b)$ is the frequency of the itemsets containing items $a$ and $b$, and $N$ is the number of observations in the dataset. If the support threshold it is too large the algorithm may fail at finding the true patterns in the dataset, whereas a small minimum support may generate an excess amount of association rules that is not fit for effective use. Confidence, for its part, is defined as:
$$
 \textnormal{Confidence}=\frac{freq(a,b)}{freq(a)}
$$
and is the frequency of itemsets that contain $a$ which also contain $b$, or how often a rule is found to be true \cite{Borgelt2005}. Consequently, the performance of these rule mining algorithms is dependent on user-specified thresholds. For both the BRL and FRL models, rules were mined using the default minimum support threshold of 10\% and confidence threshold of 80\% that is commonly used in the literature. Different thresholds on minimum support and confidence were explored but model performance did not improve thus the default parameters were retained. 

Finally, the BRL and FRL algorithms contain additional hyperparameters to specify the maximum rule cardinality and expected rule list length. The maximum rule cardinality refers to the length of the rule (or itemset). This value is typically set to 2 or 3 as it may be harder to reconcile a combination of features as being clinically logical when interpreting high cardinality rules. The prior expected list length denotes the expected number of rules in the list (excluding the null rule).

\subsubsection{Bayesian Rule Lists}

BRLs \cite{Letham2015} are used for classification problems where the goal is to learn $P(Y=1 | X)$. $Y$ is binary, and in the case of predicting brain tumour survival greater than a year, $Y = 1$ would indicate survival greater than a year and $X$ would represent a patient's features. The conditional probability distribution is represented as a decision list consisting of a series of decision rules. 

Frequent patterns (or itemsets) are first identified from the dataset using the rule mining algorithm FP-Growth \cite{Borgelt2005}. Following association rule mining, the BRL algorithm creates a posterior distribution over rule lists, given the observed data and prior assumptions (i.e., max rule cardinality and list length). These priors are used to specify rule cardinality and rule list length. Using a generative model, an initial decision list is selected and iteratively modified using Markov chain Monte Carlo sampling \cite{hastings1970monte} to generate many samples of decision lists from the posterior distribution (see Letham et al., \cite{Letham2015} for technical details). This procedure ensures the production of a variety of lists that are not dependent on one initial decision list. 

Given this posterior distribution of decision lists, new observations are classified using a point estimate (a single decision list) or the posterior predictive distribution (multiple decision lists). The point estimate is chosen as the list with the highest posterior probability from all the samples with posterior mean list length and posterior mean average rule cardinality. This estimate is called a \textit{BRL-point} \cite{Letham2015}. 

\subsubsection{Falling Rule Lists}

A FRL is an ordered decision list whereby the estimated probability of success, or  $P(Y=1 \mid X)$, monotonically decreases down the list \cite{Wang2015}. Analogous to BRL, pre-mined rules are first extracted from the data, then Bayesian modelling is used to produce a decision list (see Wang and Rudin \cite{Wang2015} for mathematical details). To approximate the FRL, Monte Carlo sampling from the posterior distribution is required to generate an initial decision list. To enforce FRL monotonicity constrains, a combination of Gibbs and Metropolis-Hastings sampling \cite{roberts1994simple} is used. Unlike BRL, following the production of an initial decision list, instead of yielding many sample lists, a point estimate is found using simulated annealing \cite{van1987simulated}.

\subsection{ML Algorithms}

Three ML classifiers were exploited to predict the 1-year survival of patients. An RF classifier \cite{breiman2001random}, which uses a multitude of decision trees \cite{quinlan1986induction} for classification, was implemented. The number of trees were selected using grid search with 3-fold cross validation. A LR classifier \cite{menard2002applied} with L2 regularisation \cite{ng2004feature} was also employed. The regularisation parameter \textit{$C_{LR}$} was selected from \{${2^{-4}, 2^{-3}, 2^{-2},...,2^{4}}$\}. Finally, an SVM classifier \cite{suykens1999least} with a radial basis function kernel \cite{park1991universal} was used. The regularisation parameter \textit{$C_{SVM}$} was selected from \{${2^{-4}, 2^{-3}, 2^{-2},...,2^{4}}$\}. Both LR and SVM penalty parameterrs were chosen using a similar range of values to the original BRL paper \cite{Letham2015}. All models were implemented using the scikit-learn package \cite{pedregosa2011scikit} and represent a group of relatively uninterpretable ML models. 
    
\section{Results}\label{results}

\subsection{Model Evaluation}

\begin{figure}[h]
    \centering
    \includegraphics[width=0.75\linewidth]{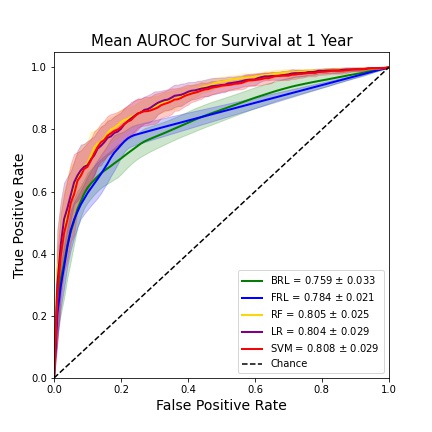}
    \caption{Mean AUROC and standard deviation for rule list and ML models are illustrated. The black dashed line represents a random classifier that performs no better than chance.}
    \label{roc_curve}
\end{figure}

The mean classification performance of the six modelling approaches on the brain tumour dataset are summarised in Table \ref{model_results}. The ROC curves for the rule lists and ML models are illustrated in Figure \ref{roc_curve}. The mean ROC curves are in bold and the standard deviation is shown by the shaded region. The AUROC could not be directly computed for the Cox PH model because it requires the probability estimates for each class which the Cox PH model does not provide. The baseline Cox PH model was outperformed by all models and FRL outperformed BRL. The rule lists were marginally outperformed by the three ML models, with SVM performing best. Notably, FRL performance was within one standard deviation of the mean performance of SVM and BRL performance was within two standard deviations. Thus the loss in BRL and FRL performance was minimal and may be mitigated by the rule list's added level of interpretability.

\begin{table}[h]
\scalebox{0.85}{
\centering
\begin{tabular}{@{}lllllll@{}}
\toprule
         & Cox         & BRL         & FRL         & RF                   & LR          & SVM         \\ \midrule
Accuracy & .711 (.045) & .758 (.028) & .782 (.021) & .807 (.024) & .805 (.030) & {.809 (.030)} \\
F1       & .698 (.045) & .754 (.031) & .780 (.021) & .804 (.025) & .802 (.030) & {.807 (.030)} \\
AUROC    & .711 (.052) & .759 (.033) & .784 (.021) & .805 (.025) & .804 (.029) & {.808 (.029)} \\ \bottomrule
\end{tabular}}
\caption{Performance metrics were assessed using nested cross-validation for three different random seeds. Mean and, in parenthesis, the standard deviation for 15 models are given. The macro-averaged F1 score is reported.}
\label{model_results}
\end{table}

\begin{figure}[h!]
    \centering
    \includegraphics[width=1\linewidth]{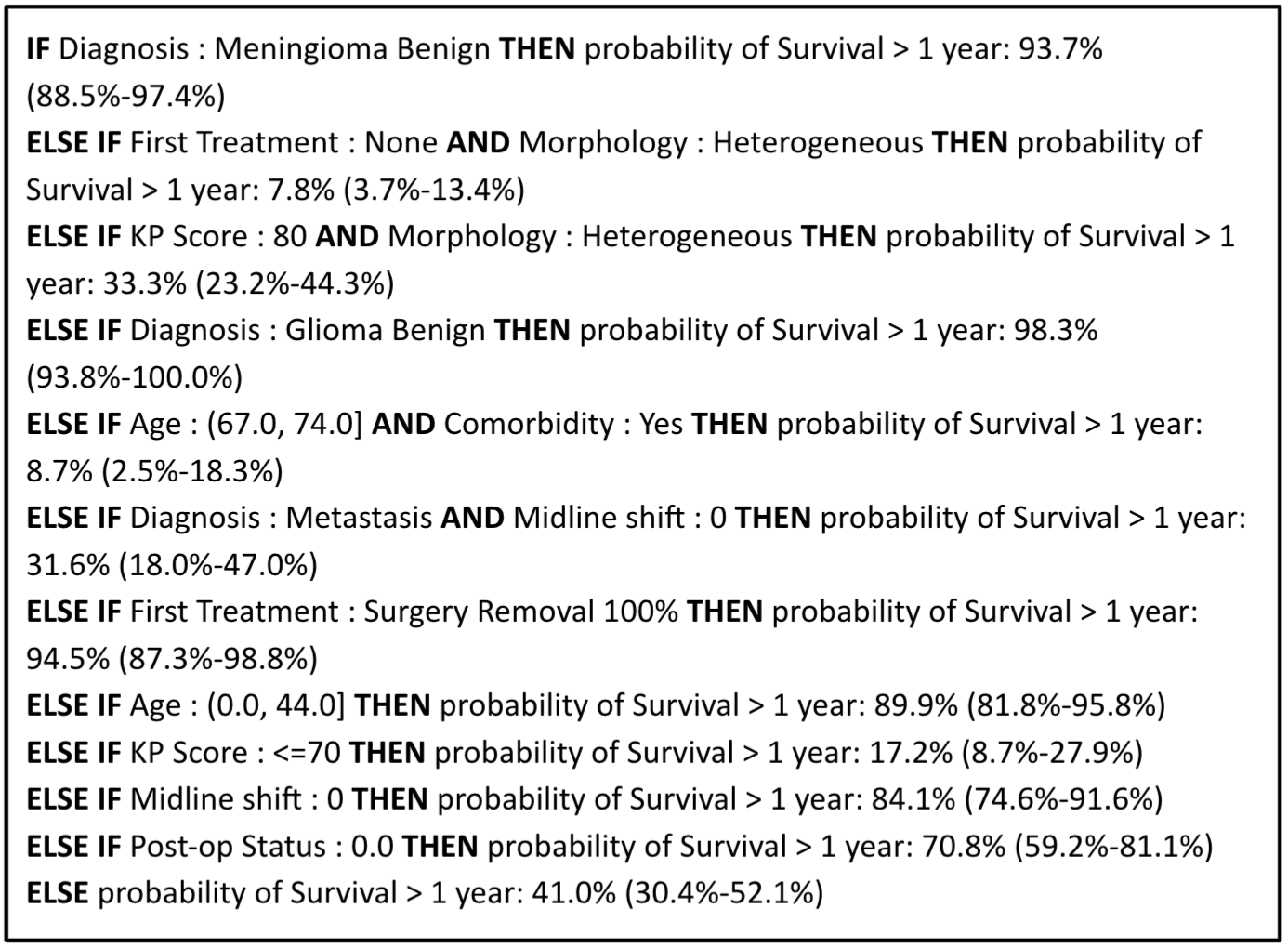}
    \caption{BRL-point estimate. The 95\% credible interval is given in parentheses.}
    \label{brl1}
\end{figure}

\begin{figure}[h!]
    \centering
    \includegraphics[width=1\linewidth]{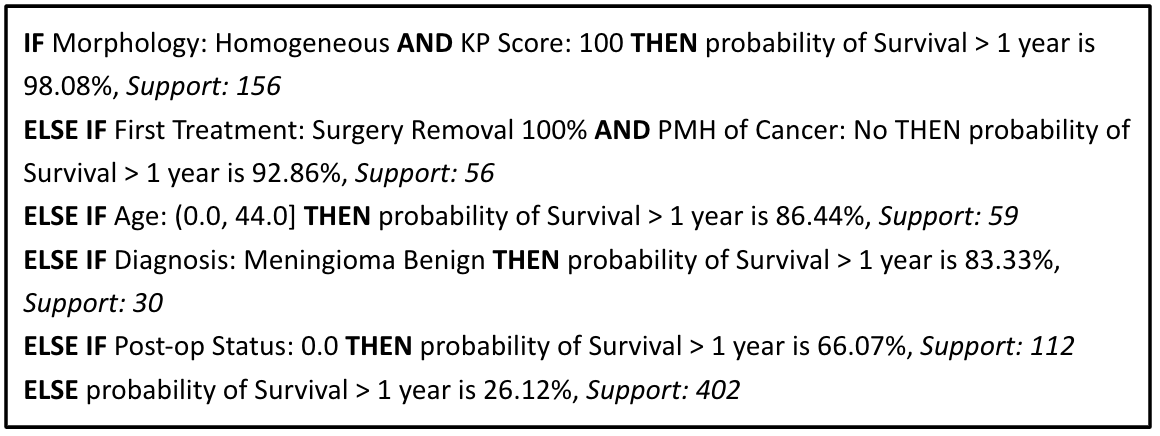}
    \caption{FRL-point estimate. The support indicates the number of patients classified by that rule.}
    \label{frl1}
\end{figure}

Figure \ref{brl1} and \ref{frl1} show point-estimates for the BRL and FRL models obtained from one of the cross validation folds. For both types of rule lists, once a patient has satisfied a rule they will not be taken into account by the cases further down the list. The final rule in the list will only consider the subset of patients that were not classified by the previous ones. The shorter decision list produced by the FRL algorithm is likely due to the model's monotonicity constraints. All rules that favour survival of less than a year are summarised into one final rule by the FRL algorithm. Comparatively, the BRL model does not follow any monotonicity constraints and so rules that favour both survival less than and greater than a year are included.

The BRL point-estimate (Figure \ref{brl1}) indicates that if a patient has a benign meningioma they will likely survive more than a year, however, if a patient receives no treatment and has a tumour with heterogeneous morphology, they will likely not survive more than a year. On the other hand, the FRL point-estimate (Figure \ref{frl1}) indicates that a tumour with homogenous morphology and a KP score of 100 are positive prognostic factors. Additionally, if a patient has a 100\% tumour resection and no previous history of cancer they are likely to survive greater than a year. By reviewing only a few rules from both lists, it is evident that features such as diagnosis, first treatment type, tumour morphology and KP score are significant survival predictors. Additionally, age of diagnosis and post-op status are repeated throughout both rule lists. 

The BRL and FRL point-estimates from the four other folds for a specific random seed are given in Appendix B, for a total of five BRL and five FRL estimates. Despite the different point-estimates, there is significant overlap in the rules. Due to the iterative construction of BRLs, there may be multiple equally good rule lists produced and it is not clear which will be returned by the model \cite{Letham2015}. The FRL model yields a single point-estimate using simulated annealing and only uses rules that favour survival greater than a year thus limiting the variability in its final decision list. The production of multiple high performing rule lists may be beneficial, as additional information on feature relationships are revealed.

\subsection{Model Interpretability}

Although the algorithms do not provide the same level of interpretability, the weighting of features at a global model level and local prediction level can be reasonably compared. Sequential feature selection was first performed on the dataset to assess feature significance. The final models interpretability was evaluated using feature importance (Cox PH, RF), post-hoc methods LIME and SHAP (RF, LR, SVM) and qualitative assessment (BRL, FRL). These aspects are discussed next.

\subsubsection{Feature Selection}

Typically, the purpose of feature selection is to remove irrelevant features or noise from the data and improve computational efficiency. We used feature selection to assess which features were most pertinent to the data and to compare these results to the feature importance of the individual models. 

Sequential feature selection from mlxtend\footnote{http://rasbt.github.io/mlxtend/use\_guide/feature\_selection/SequentialFeatureSelector/} allows for a range of \textit{k}-features to be specified and the feature combination that scores the best during cross validation is returned. Although SVM was the best performing model, the algorithm cannot be run in conjunction with the feature selector as a result of the kernel transformation, thus RF, the second best model, was run with the feature selector to provide a baseline for feature comparison. The RF default parameters were used and features were assessed using AUROC with 5-fold cross validation. Due to one-hot encoding, a total of 94 feature types were evaluated and the selector returned the 10 best features (see Table \ref{feat_selection}).

The selected features included diagnosis, age at diagnosis, SIMD score, KP score, morphology, post-op status and urgency of referral. Notably, many of these features have been previously mentioned in the literature as important prognostic variables \cite{gittleman2018survivorship, lapointe2018primary, dehcordi2012survival, davis1999conditional, stark2011surgical}, however SIMD, a social measure of deprivation in Scotland, is often associated with survival in a population study setting rather than a hospital-based setting. Nonetheless, an association between socio-economic status and cancer survival has long been demonstrated in research \cite{woods2006origins}. Despite the UK's universal healthcare, several cancer studies have linked deprivation with poorer survival outcome \cite{rachet2010socioeconomic, proctor2011inflammation, smith2014determinants}. In our dataset, 62\% of patients with a SIMD score of 4 or 5 (least deprived) survived greater than a year compared to 53\% of patients with a SIMD score of 1 or 2 (most deprived). Additionally, post-op performance status (scale 0 - 5) accounts for two of the top ten feature values returned suggesting it is valuable predictor (see Table \ref{postop_score} in Appendix for post-op performance scale). Post-op performance status assumes the patient has had surgery and a status of 0 or 1 indicates a person is fully active or restricted only with strenuous physical activity following surgery. Comparatively, a status of 4 or 5 indicates a person is completely disabled or dead, respectively. In our dataset, 70\% of patients with a post-op performance score of 0 or 1 survive more than a year compared to 37\% of patients with a status of 2-5. The majority of the 10 features selected were used by at least one of the trained models.

\begin{table}[h]
\centering
\scalebox{0.95}{
\begin{tabular}{@{}ll@{}}
\toprule
Rank & Feature                              \\ \midrule
1    & Diagnosis: Glioma Benign                   \\
2    & Diagnosis: Meningioma Benign           \\
3    & Diagnosis: Rare Tumour Benign  \\
4    & Age: (0.0, 44.0]    \\
5    & SIMD: 1.0   \\
6    & KP Score $\leq$ 70                   \\
7    & Morphology: Heterogeneous     \\
8    & Post-op Performance Status: 0.0         \\
9    & Post-op Performance Status: 1.0         \\
10   & Urgency of Referral: Suspicion of Cancer (within 2 weeks) \\ \bottomrule
\end{tabular}}
\caption{Top 10 features returned using sequential feature selector.}
\label{feat_selection}
\end{table}

\subsubsection{Feature Importance} \label{feat_imp}

An examination of a model's feature weightings can give rise to simple interpretations of how a classification is made. Although not fully transparent, this method allows moderate insight into how a model works and may assist clinicians in understanding causal factors for patient survival. The Cox PH model provides means for feature importance analysis. Notably, the Cox PH model is seldomly used to determine feature importance, rather it is used as a modelling technique to help adjust for confounders while assessing the effect side of a variable of interest. However, in the present case, the Cox PH model serves as an additional resource for feature comparison. Additionally, although SVM is the best performing model, the algorithm employs a kernel transformation hence feature importance cannot be directly computed. Thus the feature importance of RF, the second best model, was assessed using permutation importance \cite{breiman2001random}.

\begin{figure}[h!]
    \centering
    \includegraphics[width=1\linewidth]{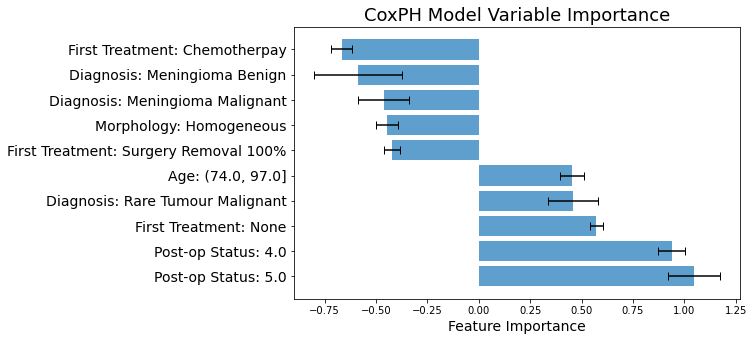}
    \caption{Bar chart showing the average Cox PH model feature importance. The coefficients were averaged across cross-validation folds and the standard deviation is shown with black error bars. The top 5 features with the largest negative coefficients and largest positive coefficients (i.e.\ the most informative features) were plotted.}
    \label{cox_feat_importance}
\end{figure}

The Cox PH model uses the variable's coefficient and standard error value to assess feature importance as illustrated in Figure~\ref{cox_feat_importance}. A feature value above 0 indicates increased risk, thus reduced survival time, whereas a value less than 0 suggests reduced risk, or increased survival time. A coefficient value of 0 indicates the feature has no importance. It is worth remarking, as discussed in Section \ref{cox_model}, the first category per feature was dropped to account for problems with high collinearity. For example, the feature \textit{Tumour Morphology} contains the values \textit{Heterogenous} and \textit{Homogenous} - \textit{Heterogenous} was dropped. Thus Cox PH model feature importance should be interpretted with caution as not all feature types are present. However, general comparisons can be made. 

According to the Cox PH model, a poor post-op status, no treatment and an older age at diagnosis are poor prognostic factors.  Comparatively, chemotherapy, homogeneous tumour morphology and 100\% tumour resection are positive prognostic factors. Similar features were used by both the rule lists and sequential feature selector. Interestingly, a malignant meningioma was a positive prognostic factor. However, the 1-year survival rate of  malignant meningioma's is around 80\% \cite{truitt2019partnership} thus highlighting the importance of the survival threshold when assessing prognostic variables.  
 
\begin{figure}[h]
    \centering
    \includegraphics[width=1\linewidth]{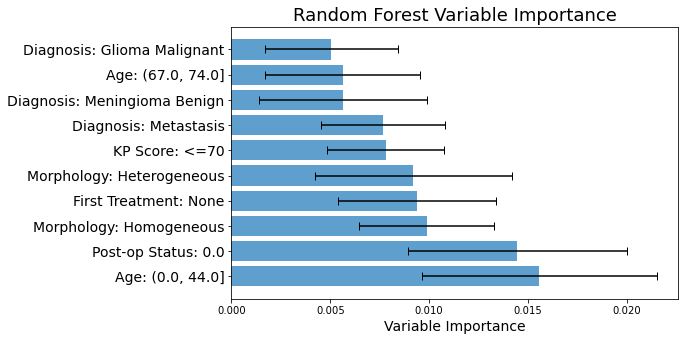}
    \caption{The 10 most influential features from the RF algorithm averaged across cross-validation folds. The standard deviation is shown with black error bars. Note that the importance does not specify positive or negative correlation.}
    \label{rf_feat_importance}
\end{figure}

RF feature importance is given in Figure \ref{rf_feat_importance}. This was obtained using permutation importance, which measures the decrease in a model's performance when a feature value is randomly shuffled \cite{breiman2001random} but does not indicate whether the feature is positively or negatively correlated with the predicted outcome. Given our dataset, which contain features with cardinalities ranging from 2 (e.g.\ Sex) to 9 (e.g.\ Diagnosis), permutation importance was preferred over the frequently-used mean decrease in impurity (MDI) \cite{breiman1996some} because the latter is often biased towards features with high cardinality \cite{strobl2007bias}. As can be seen, the three most influential features for the RF model were a younger age at diagnosis, a post-op status of 0 and heterogeneous tumour morphology. All three features were also used by the rule lists and sequential feature selector. Note the three feature values were not used to train the Cox PH model (due to collinearity problems as discussed previously), however the Cox PH model found the inverse feature values (i.e.\ an older age at diagnosis, a post-op status of 4/5 and homogeneous tumour morphology) to be informative.

Notably, the Cox PH model is seldomly used to determine feature importance, rather it is a modelling technique to help adjust for confounders while assessing the effect side of a variable of interest. 

\subsubsection{Local Surrogate Model} \label{posthoc}

\begin{figure}[htp]

\subfloat[Random Forest (RF) feature importance for a given test instance determined by LIME. The observation was correctly classified.]{%
  \includegraphics[clip,width=\columnwidth]{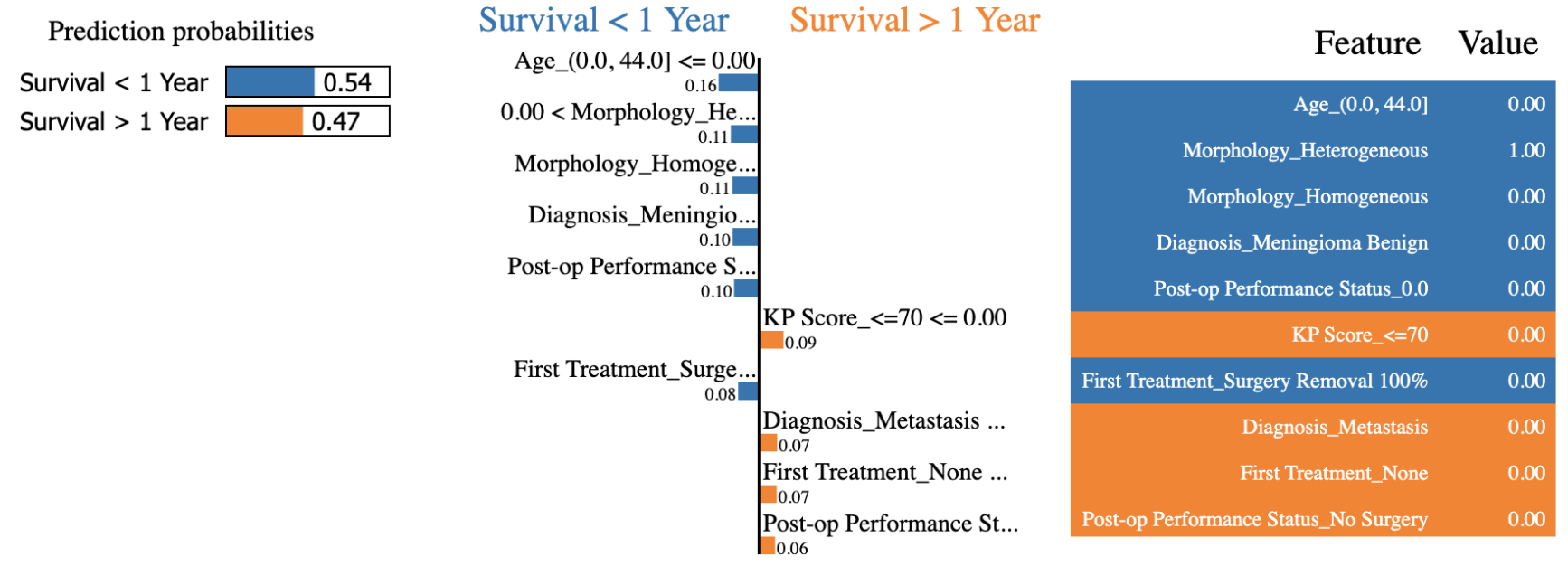}%
}

\subfloat[Logistic Regression (LR) feature importance for a given test instance determined by LIME. The observation was incorrectly classified.]{%
  \includegraphics[clip,width=\columnwidth]{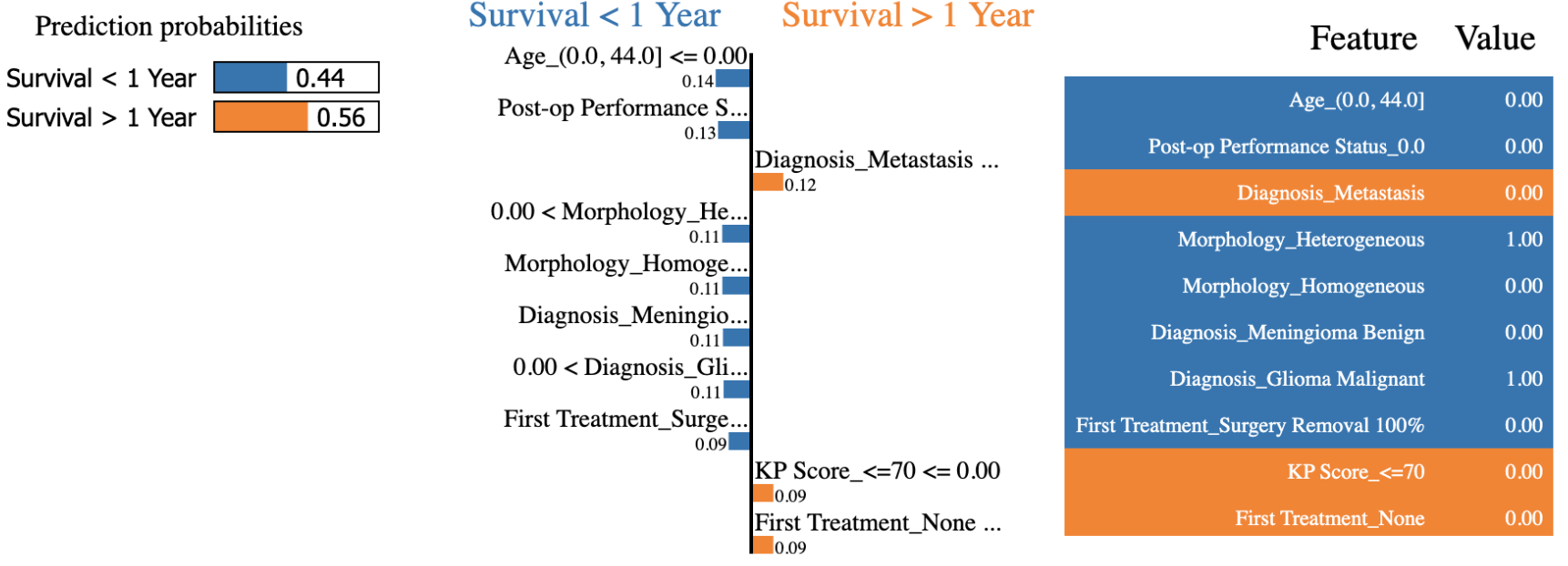}%
}

\subfloat[Support Vector Machine (SVM) feature importance for a given test instance determined by LIME. The observation was incorrectly classified.]{%
  \includegraphics[clip,width=\columnwidth]{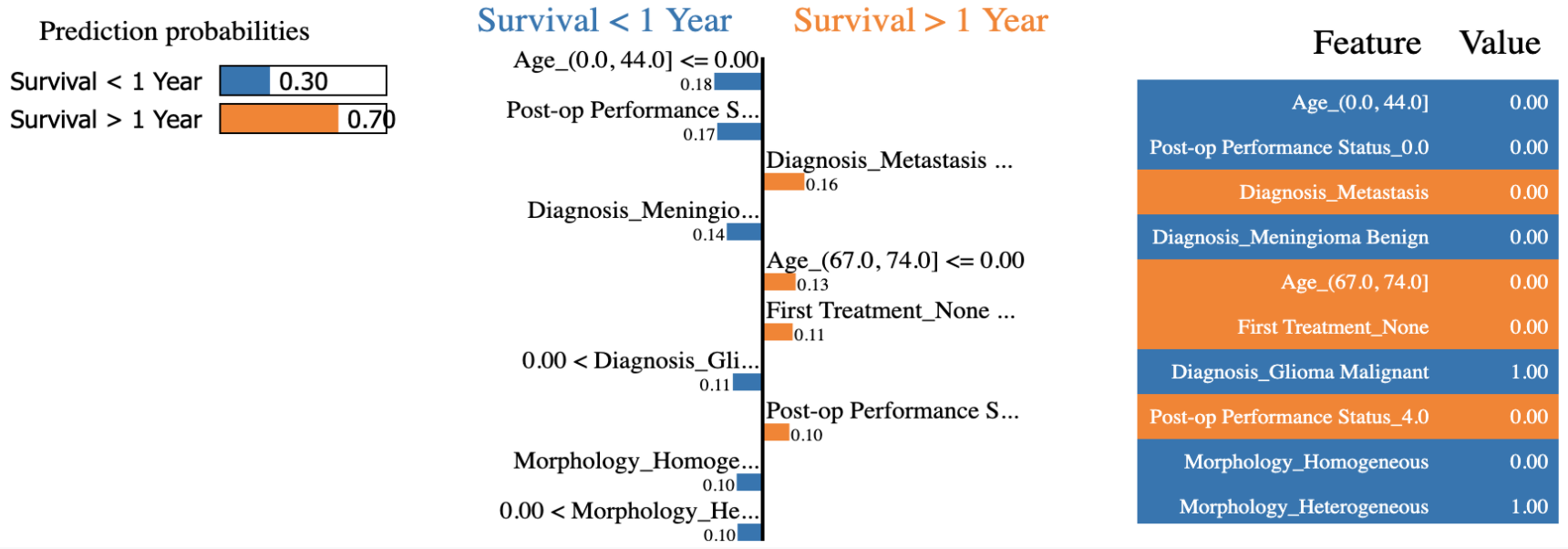}%
}

\caption{Negative (blue) features indicate survival less than a year, and positive (orange) features indicate survival greater than a year. The top 10 most influential features for the specific test instance are shown. The weight of each feature (centre image) is used to calculate the prediction probability.}
\label{LIME}
\end{figure}

\begin{figure}[!h]
    \centering
    \includegraphics[width=0.9\linewidth]{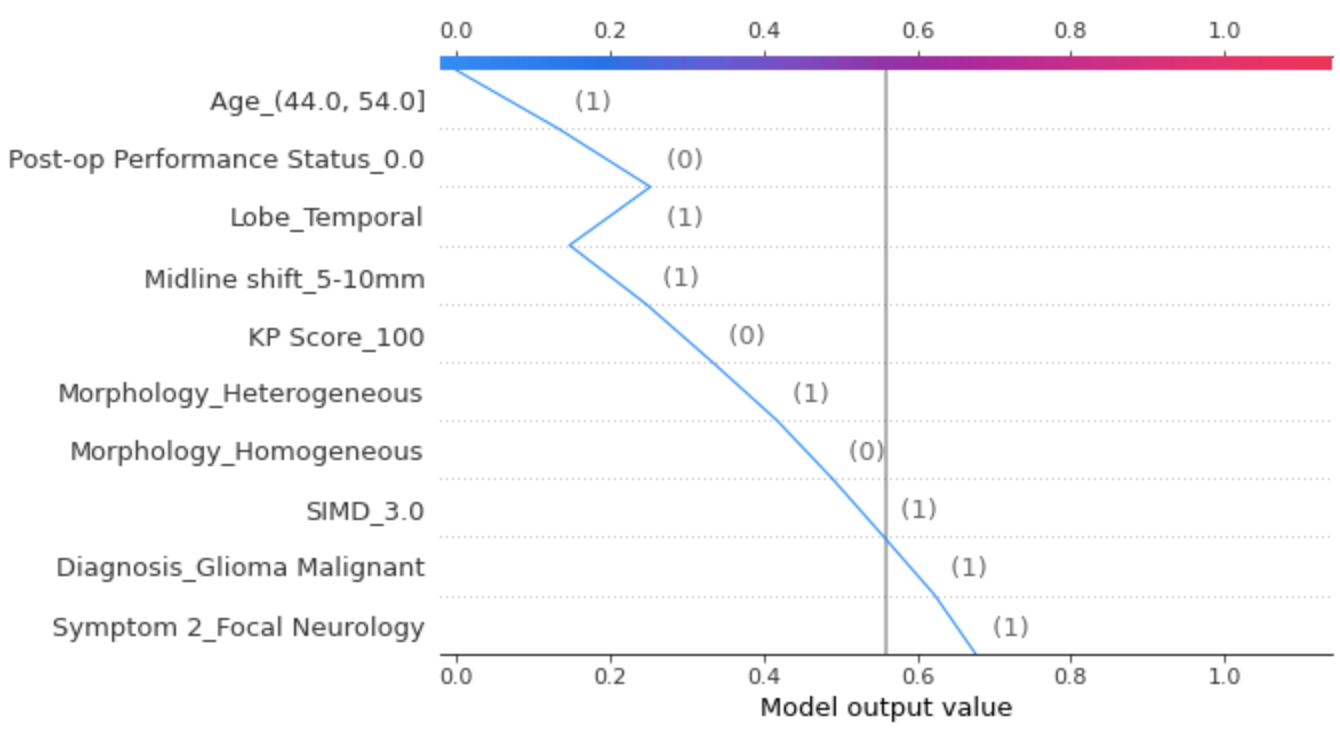}
    \caption{RF feature importance determined by SHAP. The top 10 most influential features for the specific test instance are shown. The prediction starts at a baseline value (0.410) which is the average of all predictions (not shown here). Moving from the bottom of the plot to the top, each Shapley value is added to the model's base value and either pushes to increase (positive value) or decrease (negative value) the prediction.}
    \label{rf_SHAP}
\end{figure}

Compared to feature importance, LIME and SHAP can assess interpretability at the local level for individual predictions, rather than at the global modular level. Local explanations may be more accurate than global explanations \cite{Molnar2019} and are beneficial for understanding why instances are classified incorrectly. For example, Figure \ref{LIME} illustrates LIME's explanation of an observation that was classified incorrectly by LR and SVM (i.e.\ wrongly predicted survival $>$ 1 year) and correctly by RF (i.e.\ rightly predicted survival $<$ 1 year). The top 10 influential features for the given test instance are shown. Figure \ref{rf_SHAP} shows an explanation by SHAP of the same observation classified by RF. Both surrogate models were applied to all three ML models, and the same prediction instance was compared across the three models (see Figures \ref{lr_shap} and \ref{svm_shap} in the Appendix for LR and SVM explanation by SHAP). Note that the three ML models used one-hot encoded data to make predictions thus both the presence (value = 1) and absence (value = 0) of a feature is used to make a prediction. Across all six model explanations (three ML models by two surrogate methods), similar features were used but the weighting of the features varied. For example, according to LIME, all three ML models placed the most importance on a younger age \textit{not} being present, and for LR and SVM this was follow by \textit{not} having a post-op status of 0 and \textit{not} having a metastatic tumour. In comparison, RF valued the presence of a heterogenous tumour, \textit{not} a homogenous tumour and \textit{not} having a benign meningioma. Additionally, according to LIME, six features were found to be the most influential across all models: age, morphology, diagnosis, post-op status, KP score and first treatment. In comparison, according to SHAP for the same test instance classified by RF (Figure \ref{rf_SHAP}), features such as lobe, midline shift, SIMD and symptom type were in the top 10 most influential features. The discrepancy between LIME and SHAP for the same model highlights the unreliability of post-hoc interpretability methods \cite{Rudin2019}.

\subsubsection{Qualitative Analysis} \label{experts}

In medicine, where human decision making governs the process of patient treatment, a survey over domain experts is a valuable measure of interpretability \cite{ruping2006learning}. The interpretability of our rule lists was assessed based on the clinical expertise of two of the current authors (M.P. and P.B.), who were provided with multiple point-estimates from both BRL and FRL models (see Figure \ref{brl1}, Figure \ref{frl1} and Appendix B for all rule lists provided). To mitigate any potential bias, the models were constructed without their input and only the final models were presented for relatively informal feedback on their potential clinical utility as a predictive model.

According to this evaluation, the combination of features used by the rule lists for survival prediction are informative and in-line with domain knowledge. Similar features were found influential across multiple rule lists including age at diagnosis, KP score, tumour morphology, and first cancer treatment. As expected, the diagnosis (tumour type) is also an important factor for survival. For example, benign gliomas and benign meningiomas indicted greater survival, both of which are highly treatable, with a 76\% \cite{claus2015survival} and 70\% \cite{mccarthy1998factors} five-year survival rate, respectively. Comparatively, brain metastases indicated poor survival and only have a 1-year survival rate of 17\% \cite{ekici2016survival}. The impact of these feature combinations have been well-established in the literature \cite{gittleman2018survivorship, lapointe2018primary, dehcordi2012survival, davis1999conditional} and were re-produced by the rule-lists. An agreement between model predictions and clinical knowledge is essential for establishing trust in the models decision making process, as well as trust that the model will make accurate predictions when applied to new data.

\subsection{Model Comparison}

\begin{figure}[h]
    \centering
    \includegraphics[width=0.85\linewidth]{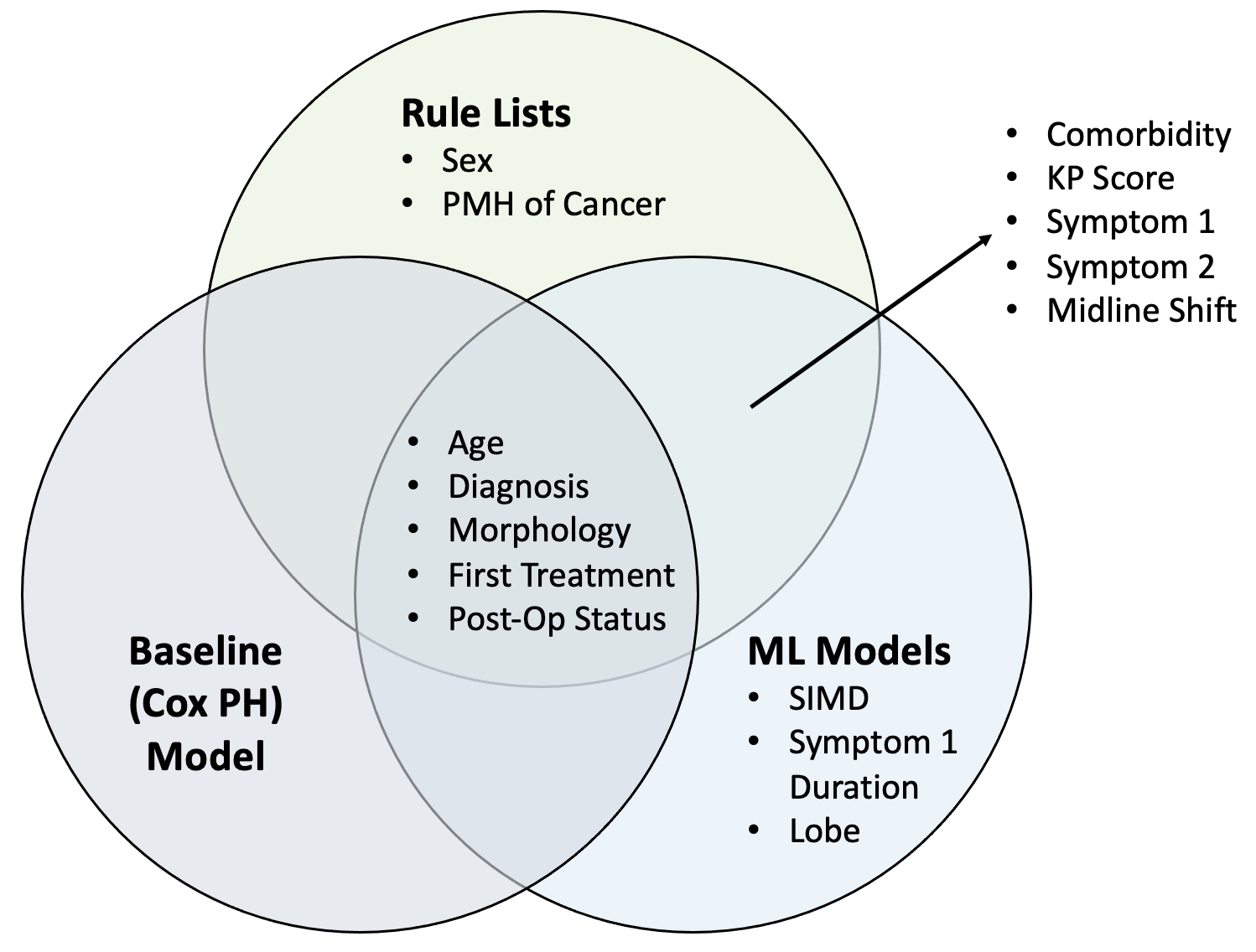}
    \caption{Venn diagram of the features used by the three main groups of models for survival prediction. Features used by all model types are found in the centre. Note the general feature types are compared between models instead of the individual feature values.}
    \label{ven_diagram}
\end{figure}

The features found important by the three model types (Cox PH, rule lists, ML) for survival prediction are summarised visually by a Venn diagram in Figure \ref{ven_diagram}. For ease of evaluation, the general feature types are used for comparison between models (e.g.\ sex) instead of the individual feature values (e.g.\ female or male). Additionally, the features used by the ML models are for a single test instance thus only general comparisons can be made. 

All of the features returned by the sequential feature selector (see Table \ref{feat_selection}), except \textit{Urgency of Referral}, were used by at least one of the three model types. All features used by the baseline Cox PH model were also found important by the rule lists and ML models. We only investigated the top 10 most influential features used by the Cox PH model hence there is likely more overlap in feature significance between the three models. The features identified as influential by all models included age, diagnosis, morphology, first treatment and post-op status. The rule lists and ML models had multiple overlapping features including comorbidity, KP Score, symptom 1, symptom 2 and midline shift. Only the rule lists found sex and history of cancer to be significant, while only the ML models found SIMD, symptom 1 duration and lobe to be important. The importance of symptomatology data is interesting, as symptoms are often talked of in regards to time to diagnosis, but less often with regards to prognosis. When looking at the RF global top 10 important features (see Figure \ref{rf_feat_importance}), symptom and sign features were not found relevant, but when using SHAP to look at a specific test instance from the RF model (see Figure \ref{rf_SHAP}), as well as LR and SVM (see Figure \ref{lr_shap} and \ref{svm_shap} in Appendix), symptom information was relevant. As mentioned above, sex was only found relevant for the rule lists which was surprising. Majority of research into sex differences and brain tumour survival have focused on glioblastomas, which find that men are more likely to develop, and die of, glioblastomas than women \cite{tian2018impact, ostrom2018females, sun2015integrative}. In comparison, women are twice as likely to develop meningiomas compared to men but no significant difference in outcome have been reported \cite{sun2015integrative, holleczek2019incidence}. Although sex may play an important role in individual tumour types, across all tumour types sex may be less influential. Finally, four features were not used by any model: Urgency of Referral, Sign 1, Side and Max Size. Note Urgency of Referral was found to be important by the sequential feature selector (see Table \ref{feat_selection}) but not by the algorithms. Given the importance of symptom 1 and symptom 2, the irrelevance of sign 1 (the first onset objective clinical abnormality) is perhaps the most surprising. Although in our dataset 42\% of patients presented with no signs, as symptoms dominate, hence this feature may be less informative for classification. Features which are not ranked as important by models should arguably be given less attention in clinical evaluation and the use of such features should be reviewed if they currently play a decision role in the assessment. It is beneficial to simplify models, and decision making, by reducing some of the noise over features, as there is often extraneous data in healthcare that can impair decision making rather than support it.

Model interpretability offers valuable insight on the importance of the individual features for survival prediction. Although there is an overlap in the features used by the models, the weighting of each feature differs and a comparison of this weighting may prove the most informative as it helps one understand what drives the model's performance. This is illustrated best with LIME and SHAP where for a given test instance the weightings of each feature used for classification are returned. For example, for the test instance discussed in Section \ref{posthoc}, where RF correctly classified the instance but LR and SVM did not, the weighting of the feature \textit{not having a metastatic tumour} varied. According to LIME, RF placed a weight of 0.09 on not having a metastatic tumour, compared to LR and SVM which gave this feature a weighting of 0.12 and 0.16, respectively. Additionally, according to SHAP, this feature was \textit{not} in the RF's top 10 informative features, but it was for LR and SVM. Hence an understanding of a features predictive power and how this varries between models is valuable for prognostic decision making.

\section{Discussion}\label{discussion}

The results summarized in Table \ref{model_results} demonstrate that the BRL and FRL algorithms were outperformed slightly by popular ML models. However, the innate interpretability of rule lists may mitigate potential performance loss, especially in health-care where model transparency is essential for its integration into- and influence on decision making \cite{Rudin2019}. The original BRL paper derived a stroke prediction model and found that the BRL-point estimate performed on par with SVM and only $\sim$2\% worse than RF (their best performing model). Comparatively, the average BRL performance for brain tumour survival prediction was $\sim$5\% worse than SVM (our best performing model). Notably the dataset used by the original authors was more than $12\times$ the size of ours, hence the additional training data may have helped BRL performance. The minimal decrease in BRL performance for both stroke and brain tumour survival classification highlights the potential of innately interpretable algorithms for high-stake predictions. 

Our results are also similar to that of Wang and Rudin \cite{Wang2015}, who found that FRL performance was slightly worse, but comparable to RF, LR and SVM on four public datasets. FRL performance on our dataset is within one standard deviation of the ML algorithms mean performance. The minimal loss in FRL performance is not surprising as the model's strong monotonicity constraints may sacrifice performance \cite{Wang2015}.

More generally, the slight loss in rule list performance compared to the ML models may be due to our dense heterogeneous dataset and limited sample size. Rule list algorithms optimise over pre-mined rules, rather than the entire feature space, hence the algorithm's computational effort scales with the number of antecedents rather than the number of features \cite{Letham2015}. Thus in general, less computation is required when the data is space. For BRL, depending on the hyperparameters, the number of antecedents used ranged from $\sim$850 to $\sim$2500 across the training folds (this information was not available for FRL). Although we attempted to reduce dataset complexity (e.g.\ by grouping rare tumours together and combining symptom and sign data into larger domains), rule list models may be better suited for a dataset with sparse features, especially given a small sample size. A model with higher statistical capability, such as the ML models or neural networks, may be better suited for handling dense features with limited training examples, however such models are often the least transparent. 

Rule-lists are a promising interpretable model but are currently limited to categorical features and binary classification (see \cite{proencca2020interpretable} for a recent proposal of a multi-class rule list algorithm). Similar to decision trees, rule lists cannot deal with linear relationships and due to their categorical input requirement, rules can only produce step-like predictions \cite{Molnar2019}. Additionally, a caveat to rule list interpretability is that such models are generally not stable. There may be multiple equally good rule lists and it is not clear which will be returned by the algorithm \cite{Letham2015}. Furthermore, small alterations in the training dataset can result in a different rule list, which is especially problematic given our small dataset size. Thus using the full posterior of rule-list samples (i.e.\ BRL-post) rather than a single rule-list (i.e. BRL-point) may improve model stability, however, the classifier would no longer be interpretable. Despite these limitations, rule-lists are easy to interpret, and only the relevant features for rule list construction are selected. The classification of a patient is fast, whereby only a few binary statements need to be reviewed. BRLs offer a more fine-grained approach than FRLs, whereby rules that favour survival less than and greater than a year are used. In comparison, FRLs have the added benefit of automatically stratifying patients by risk thus a clinician only needs to look at the top few rules to classify the most high risk patients. Similar to Senders et al. \cite{senders2020online}, an interactive graphical representation of the rule list model could be created, whereby a clinician inputs a patient's features and a prediction is returned along with the decision rules used to make that prediction. Finally, the integration of additional clinical information such as blood tests or genetic data may improve the clinical validity of rule list models. Blood tests are now being investigated as a means for brain tumour diagnosis \cite{gray2018health, podnar2019diagnosing} and genetic alterations have shown to be effective predictors for tumour prognosis \cite{Fulop2019predicting, molinaro2019genetic}.

Although the innate interpretability of rule-lists cannot be directly compared to the post-hoc interpretability of ML algorithms, general conclusions can be made. Traditional variable importance algorithms are limited to group-level analysis but LIME and SHAP enable the explanation of individual predictions to understand the contribution of each predictor. Our results showed that similar features were used by the three ML models for making predictions. However, the weighting of each feature varied between models and across interpretability techniques. The different feature weightings between models may explain the variability in model performance, but the difference within interpretability techniques highlights the inconsistency between post-hoc methods. Because post-hoc methods assess interpretability after model construction, explanations can prove misleading and unreliable \cite{Rudin2019}, and there is a growing body of literature that has questioned the credibility of LIME and SHAP \cite{Rudin2019, laugel2019dangers, slack2020fooling, dimanov2020you, lee2019developing}. There is a large toolbox of post-hoc methods \cite{Molnar2019} available to assess model interpretability, however, the absence of a cohesive definition for interpretability makes it difficult to assess the quality of different interpretability techniques. By agreeing on clear quantitative metrics for interpretability, the development of robust trusted interpretability techniques can follow \cite{elshawi2020interpretability}.

\subsection{Strengths and Limitations}\label{limitations}

This was the first study to apply classification algorithms to a novel brain tumour dataset for 1-year survival prediction. Brain tumours are a rare but deadly disease, and compounded by their heterogeneity, an accurate prognosis by clinicians is a challenging task. Also, to our knowledge, this is the first study to compare intrinsic and post-hoc interpretability methods for the assessment of predictive brain tumour survival models. As ML continues to advance, the development of tools to assist clinicians with these high-stake medical decisions by providing trustworthy data-driven support will become essential for acceptability. As we have already discussed, some of the advantages of rule lists in clinical practice are as follows: their innate interpretability, their simple \textit{if...then...} structure is easy to follow and predictions with rules are fast (only a few binary statements need to be assessed). 

Nonetheless, several important limitations remain. In the present study, only clinical prognostic factors were used for the prediction of survival. As previously mentioned, blood tests and molecular genetic alterations have been recognised as powerful prognostic and predictive markers in brain tumour survival \cite{gray2018health, podnar2019diagnosing, molinaro2019genetic, eckel2015glioma, hartmann2010patients, diplas2018genomic}. Hence the integration of the different data types may improve current survival predictions. However, an increase in data dimensionality is an important consideration especially when rule lists are employed. In addition, the data used in this study was heterogeneous and a second validation dataset is required for confirmation of our results. With the addition of new data, this study could also be extended to explore the application of post-hoc interpretability methods to neural networks. As discussed above, rule-lists currently require categorical features and are limited to binary classification (see \cite{proencca2020interpretable} for multi-class rule list algorithm proposal). Extension of the rule list algorithm for multi-class classification or regression is an important next step for improving rule-list performance and constructing a competitive interpretable rule list classifier. 

Finally, although not directly related to the ML models, our dataset required imputation and discretisation. There is the potential for imputation to introduce bias into the data \cite{sterne2009multiple, jakobsen2017and} and the chosen imputation method can influence the final results \cite{stavseth2019handling}. Additionally, although discretisation can be used to improve the clarity of classification models by extracting useful feature intervals, the split of the feature will also effect a model's performance. Both pre-processing techniques have the potential to reduce model performance which we sought to mitigate during dataset preprocessing through the exploration of multiple imputation and discretisation algorithms. 

\section{Conclusion}\label{conclusion}

This study investigated the performance and interpretability of multiple algorithms for the prediction of brain tumour survival on a novel dataset. We have demonstrated that rule list algorithms create reliable and understandable results that have clinical relevance without significant compromise in model performance. Rule lists and other interpretable models may provide an advantage over traditional clinician assessment of prognosis by weighting potential risk factors and stratifying patients accordingly. Additionally, the slight superiority in ML model performance may be less important than the agreement of features between different models within a clinical context since the latter can provide more confidence as to the importance of those features. As shown in the current work, the reliability of LIME and SHAP for assessing feature importance is questionable and the methods are vulnerable to failures. Interpretability is crucial for the implementation of ML algorithms in healthcare, because prediction tools inform, but do not single-handedly direct, clinical decision making. A model's ability to explain its predictions is essential for establishing a user's trust in the model. The interpretable models introduced in this work attempt to bridge the gap between ML research and integration into clinical practice. Rule lists are not meant to be a direct competitor for black box classifiers, but rather a useful tool that can assist humans with high-stake decisions by providing trustworthy data-driven support. Further clinical utility in the current context may come from using other interpretable approaches with clinical and molecular (or imaging) data, where it is difficult for a clinician to determine what the most informative features are across the different data sources. Interpretable models are a natural choice for the domain of predictive medicine, and whether the model is innately interpretable or post-hoc methods are utilised, the validation and integration of such models into clinical practice is an important next step for improving patient outcomes in a trusted way.

\section*{Funding}

MTCP is supported by Cancer Research UK Brain Tumour Centre of Excellence Award (C157/A27589).

\section*{Conflict of interest statement}
The authors declare no conflict of interest.

\newpage

\section*{Appendix A}\label{app_a}
\subsection{Glossary of Dataset Features}

{\footnotesize
\begin{longtable}{p{2cm}p{4.2cm}p{3.3cm}p{1.3cm}}

\caption{Overview of dataset variables including their descriptions, value and percentage of each value present in the final dataset following imputation and discretisation.} \label{tab:long} \\

\toprule
Name & Description & Value & Proportion (\%) \\* \midrule
\endfirsthead

\multicolumn{4}{c}%
{{\bfseries \tablename\ \thetable{} -- continued from previous page}} \\
\toprule
Name & Description & Value & Proportion (\%) \\* \midrule
\endhead

\multicolumn{4}{r}{{Continued on next page}}\\ \bottomrule
\endfoot
\bottomrule
\endlastfoot

\multirow{6}{*}{Age} & \multirow{6}{*}{the age of a patient} 
 & 0-44 & 17.1 \\
 &  & 45-54 & 18.7 \\
 &  & 55-61 & 16.0 \\
 &  & 62-67 & 15.8 \\
 &  & 68-74 & 16.9 \\
 &  & 75$+$ & 15.5 \\* \midrule
\multirow{2}{*}{Sex} & \multirow{2}{*}{the sex of the patient} & Male & 50.5 \\
 &  & Female & 49.5 \\* \midrule
\multirow{2}{*}{\begin{tabular}[c]{@{}l@{}}History of \\ Cancer \end{tabular}} & \multirow{2}{*}{\begin{tabular}[c]{@{}l@{}}whether the patient has a \\ past medical history of cancer\end{tabular}} & Yes & 18.2 \\
 &  & No & 81.8 \\* \midrule
\multirow{2}{*}{Comorbidity} & \multirow{2}{*}{\begin{tabular}[c]{@{}l@{}}the presence of another illness \\ or disease occurring in a patient\end{tabular}} & Yes & 47.7 \\
 &  & No & 52.3 \\* \midrule
\multirow{5}{*}{\begin{tabular}[c]{@{}l@{}}Scottish Index \\ of Multiple \\ Deprivation \\ (SIMD)\end{tabular}} & \multirow{5}{*}{\begin{tabular}[c]{@{}l@{}}a measure of deprivation of \\  the area a patient lives \\ from most deprived (ranked 1) \\ to least deprived (ranked 5)\end{tabular}} & 1 & 13.9 \\
 &  & 2 & 22.6 \\
 &  & 3 & 21.0 \\
 &  & 4 & 18.8 \\
 &  & 5 & 23.7 \\* \midrule
\multirow{4}{*}{\begin{tabular}[c]{@{}l@{}}Karnofsky \\ Performance \\ Score \\ (KP Score)\end{tabular}} & \multirow{4}{*}{\begin{tabular}[c]{@{}l@{}}a common measure in oncology \\ to assess the functional state \\ of a patient\end{tabular}} & 100 & 37.6 \\
 &  & 90 & 28.6 \\
 &  & 80 & 14.6 \\
 &  & $\leq$70 & 19.2 \\* \bottomrule
 
\pagebreak
  
\multirow{6}{*}{Symptom 1} & \multirow{6}{*}{\begin{tabular}[c]{@{}l@{}}the first symptom type a \\  patient presented with \\ (reported by the patient)\end{tabular}} & Focal Neurology & 34.6 \\
 &  & Headache & 28.4 \\
 &  & Fits/Faints/Falls & 17.1 \\
 &  & Behavioural/Cognitive & 16.7 \\
 &  & Other/Non-specific & 2.4 \\
 &  & Non-specific Neurological & 0.8 \\* \midrule
 
\multirow{5}{*}{\begin{tabular}[c]{@{}l@{}}Symptom 1 \\ Duration\end{tabular}} & \multirow{5}{*}{\begin{tabular}[c]{@{}l@{}}the length of time of a patient's \\ first symptom)\end{tabular}} 
& 0-2 weeks & 20.6 \\
 &  & 3-4 weeks & 20.1 \\
 &  & 5-8 weeks & 19.5 \\
 &  & 9-20 weeks & 20.4 \\
 &  & 20-52 weeks & 19.4 \\* \midrule
 
\multirow{6}{*}{Symptom 2} & \multirow{6}{*}{\begin{tabular}[c]{@{}l@{}}the second symptom type a \\  patient presented with \\ (reported by the patient)\end{tabular}} & Focal Neurology & 31.3 \\
 &  & No Symptoms & 30.4 \\
 &  & Behavioural/Cognitive & 18.9 \\
 &  & Fits/Faints/Falls & 9.1 \\
 &  & Headache & 6.4 \\
 &  & Other/Non-specific & 3.9 \\*  \midrule
 \multirow{6}{*}{Sign 1} & \multirow{6}{*}{\begin{tabular}[c]{@{}l@{}}the first sign type a patient \\ presented with \\ (observed by the physician)\end{tabular}} & No Signs & 42.7 \\
 &  & Neurological & 36.2 \\
 &  & Cognitive & 15.0 \\
 &  & Cranial Nerve & 5.0 \\
 &  & Other & 0.8 \\
 &  & Behavioural & 0.3 \\* \midrule
\multirow{4}{*}{\begin{tabular}[c]{@{}l@{}}Urgency of \\ Referral\end{tabular}}& \multirow{4}{*}{\begin{tabular}[c]{@{}l@{}}the patient's urgency of \\ referral from primary care\end{tabular}} & Emergency & 59.7 \\
 &  & \begin{tabular}[c]{@{}l@{}}Suspicion of Cancer \\ (within 2 weeks)\end{tabular} & 17.3 \\
 &  & Soon (up to 3-4 weeks) & 2.9 \\
 &  & Routine (up to 12 weeks) & 20.1 \\* \bottomrule

 \pagebreak
 
\multirow{9}{*}{\begin{tabular}[c]{@{}l@{}}Diagnosis \\ (or Tumour \\ Type)\end{tabular}} & \multirow{9}{*}{\begin{tabular}[c]{@{}l@{}}the type of brain tumour a \\ patient was diagnosed with\end{tabular}} & Glioma Malignant & 46.5 \\
 &  & Metastasis & 19.0 \\
 &  & Meningioma Benign & 13.6 \\
 &  & Glioma Benign & 7.1 \\
 &  & Rare Tumour Benign & 4.7 \\
 &  & Lymphoma Malignant & 4.1 \\
 &  & Meningioma Malignant & 2.3 \\
 &  & Rare Tumour Malignant & 1.5 \\
 &  & Hemangioblastoma Benign & 1.2 \\* \midrule
 
\multirow{4}{*}{Max Size} & \multirow{4}{*}{a measure of the tumour size} & $\leq$ 20 & 19.7 \\
 &  & 21-40 & 38.1 \\
 &  & 41-60 & 30.8 \\ 
 &  & $\geq$ 61 & 11.4 \\* \midrule
\multirow{4}{*}{Side} & 
\multirow{4}{*}{\begin{tabular}[c]{@{}l@{}}the side of the brain the \\ tumour is located\end{tabular}}
& Left & 41.9 \\
 &  & Right & 41.2 \\
 &  & Both Left and Right & 11.6 \\
 &  & Midline & 5.3 \\* \midrule
 
\multirow{7}{*}{Lobe} & 
\multirow{7}{*}{\begin{tabular}[c]{@{}l@{}}the lobe where the tumour \\ is located\end{tabular}} & Frontal & 34.2 \\
 &  & Temporal & 21.6 \\
 &  & Parietal & 14.6 \\
 &  & Multiple & 12.2 \\
 &  & Cerebellar & 7.3 \\
 &  & Brainstem & 5.7 \\
 &  & Occipital & 4.4 \\* \midrule

\multirow{3}{*}{Morphology} & \multirow{3}{*}{\begin{tabular}[c]{@{}l@{}}the histological classification \\ of the tumour based on the \\ cell types present\end{tabular}} & Heterogenous & 68.5 \\
 &  & Homogenous & 31.5 \\
  &  &  & \\* \bottomrule
  
\pagebreak
 
\multirow{4}{*}{Midline Shift} & \multirow{4}{*}{\begin{tabular}[c]{@{}l@{}}a measure of the tumour's \\ horizontal shift from the mid \\ (centre) line\end{tabular}} & 0 & 43.3 \\
 &  & $<$ 5mm & 28.1 \\
 &  & 5-10mm & 17.4 \\
 &  & $>$ 10mm & 11.2 \\* \midrule
 
 \multirow{9}{*}{\begin{tabular}[c]{@{}l@{}} First Treatment \end{tabular}} & \multirow{9}{*}{\begin{tabular}[c]{@{}l@{}}the type of first cancer \\ treatment \end{tabular}} & Surgery Removal 100\% & 16.0 \\
 &  & Surgery Removal 90-99\% & 24.4 \\
 &  & Surgery Removal 50-89\% & 6.4 \\
 &  & Surgery Removal $<$50\% & 4.9 \\
 &  & Biopsy & 16.9 \\
 &  & Radiotherpay & 5.5 \\
 &  & Chemotheapy & 0.9 \\
 &  & Other (e.g.\ steroids) & 2.5 \\
 &  & No Treatment & 22.5 \\* \midrule
 
\multirow{7}{*}{\begin{tabular}[c]{@{}l@{}}Post-operative \\ Performance \\ Status\end{tabular}} & \multirow{7}{*}{\begin{tabular}[c]{@{}l@{}}a measure of a patient’s level \\ of functioning following surgery \\ in terms of their ability for \\ self-care, daily activity, and \\ physical ability\end{tabular}} & 0 & 31.5 \\
 &  & 1 & 27.4 \\
 &  & 2 & 6.2 \\
 &  & 3 & 1.9 \\
 &  & 4 & 1.4 \\
 &  & 5 & 0.2 \\
 &  & No Surgery & 31.4 \\* \bottomrule
\end{longtable}}

\clearpage


\subsection{Preprocessing}

\begin{figure}[h]
    \centering
    \includegraphics[width=1\linewidth]{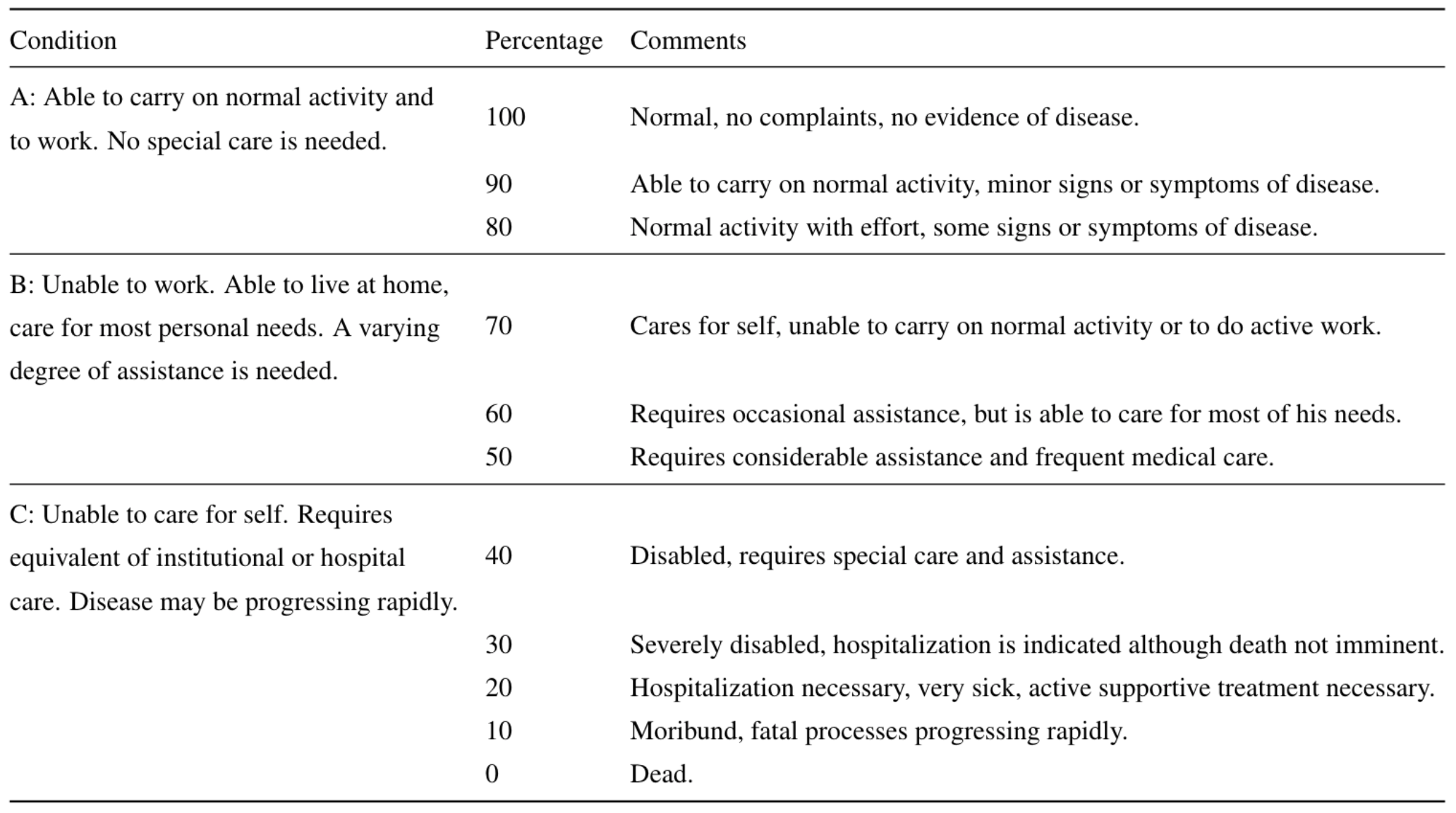}
    \caption{The original description of the Karnofsky performance status given by Karnofsky and Burchenal \cite{karnofsky1948use}.}
    \label{kps_table}
\end{figure}

\begin{table}[h]
\centering
\scalebox{0.9}{
\begin{tabular}{@{}lll@{}}
\toprule
Group & Symptom Domain            & Symptom Examples                          \\ \midrule
1     & Headache                  & Headache                                  \\
2     & Behavioral/Cognitive      & Confusion, memory loss, strange behaviour \\
3     & Focal Neurology           & Ataxia, vertigo, vision problems,         \\
4     & Fits, faints or falls     & Seizure, collapse, convulsion             \\
5     & Non-specific neurological & Poor balance, dizziness, gait abnormality \\
6     & Other/non-specific        & Vomiting, lethargy, sweating              \\ \bottomrule
\end{tabular}}
\caption{Symptom domain classifications based on Ozawa et al. \cite{Ozawa2019},  with examples of symptom types in the brain tumour dataset.}
\label{symp_domain}
\end{table}

\begin{table}[h]
\centering
\resizebox{\textwidth}{!}{%
\begin{tabular}{@{}lll@{}}
\toprule
Group & Sign Domain   & Sign Examples                                                                                                                                                                                                                                                                                                                                                                                                                                                                                                                \\ \midrule
1     & No signs      & No signs                                                                                                                                                                                                                                                                                                                                                                                                                                                                                                                     \\ \midrule
2     & Behavioral    & \begin{tabular}[c]{@{}l@{}}Behaviour signs anxiety (e.g. fast speech, tremor, voices anxiety, crying)\\ Behaviour signs depression (e.g. voices low mood, crying)\\ Behaviour (withdrawn/apathetic) - not depressed\\ Behaviour (aggressive/paranoid) - not anxious\end{tabular}                                                                                                                                                                                                                                             \\ \midrule
3     & Cognitive     & \begin{tabular}[c]{@{}l@{}}Cognitive - problems performing tasks (e.g. calculation, planning, VF)\\ Cognitive - problems with memory (forgetfulness)\\ Cognitive - reduced conscious level/drowsiness (reduced GCS)\\ Cognitive - other non-specific confusion\end{tabular}                                                                                                                                                                                                                                                  \\ \midrule
4     & Neurological  & \begin{tabular}[c]{@{}l@{}}Dysphasia - Receptive\\ Dysphasia - Expressive\\ Dysarthria - slurred or slow or staccato\\ Unilateral weakness (UMN type \textgreater{}=2 of arm/leg/face)\\ Unilateral numbness (\textgreater{}=2 of arm/leg/face, or spinothalamic type)\\ Problems with dexterity/fine manipulation\\ Problems walking/unsteadiness (weakness/numbness)\\ Problems walking/ataxia\\ Problems with visual acuity (unilateral or bilateral)\\ Problems with visual field (unilateral or bilateral)\end{tabular} \\ \midrule
5     & Cranial Nerve & \begin{tabular}[c]{@{}l@{}}Papilloedema\\ Diplopia CN problems 3, 4 or 6\\ Nystagmus (unilateral or bilateral)\\ Facial numbness/tongue numbness (CN 5)\\ Facial weakness (CN 7)\\ Reduced smell/taste (CN 1 or 7)\\ Deafness (unilateral/bilateral) (CN 8)\\ Problems swallowing (dysphagia) (CN 9, 10)\\ Problems with volume of speech (dysphonia) (CN 10)\end{tabular}                                                                                                                                                   \\ \midrule
6     & Other         & Other                                                                                                                                                                                                                                                                                                                                                                                                                                                                                                                        \\ \bottomrule
\end{tabular}%
}
\caption{Sign domain classifications based on clinical expertise of some of the current authors. All examples are from the Brain Tumour dataset.}
\label{sign_grouping}
\end{table}

\begin{table}[h]
\centering
\resizebox{\textwidth}{!}{%
\begin{tabular}{@{}ll@{}}
\toprule
Grade & Description                                                                                                                                                                                           \\ \midrule
0     & Fully active, able to carry on all pre-disease performance without restriction.                                                                                                                       \\
1     & \begin{tabular}[c]{@{}l@{}}Restricted in physically strenuous activity but ambulatory and able to carry \\ out work of a light or sedentary nature, e.g., light house work, office work.\end{tabular} \\
2     & \begin{tabular}[c]{@{}l@{}}Ambulatory and capable of all self-care but unable to carry out any work \\ activities; up and about more than 50\% of waking hours.\end{tabular}                           \\
3     & \begin{tabular}[c]{@{}l@{}}Capable of only limited self-care; confined to bed or chair more than 50\% of \\ waking hours.\end{tabular}                                                                 \\
4     & \begin{tabular}[c]{@{}l@{}}Completely disabled; cannot carry on any self-care; totally confined to bed \\ or chair.\end{tabular}                                                                       \\
5     & Dead.                                                                                                                                                                                                 \\ \bottomrule
\end{tabular}%
}
\caption{Description of a patient's performance status (or functional state) developed by the Eastern Cooperative Oncology Group \cite{oken1982toxicity}.}
\label{postop_score}
\end{table}

\clearpage


\subsection{Local Surrogate Model}

\begin{figure}[h!]
    \centering
    \includegraphics[width=1\linewidth]{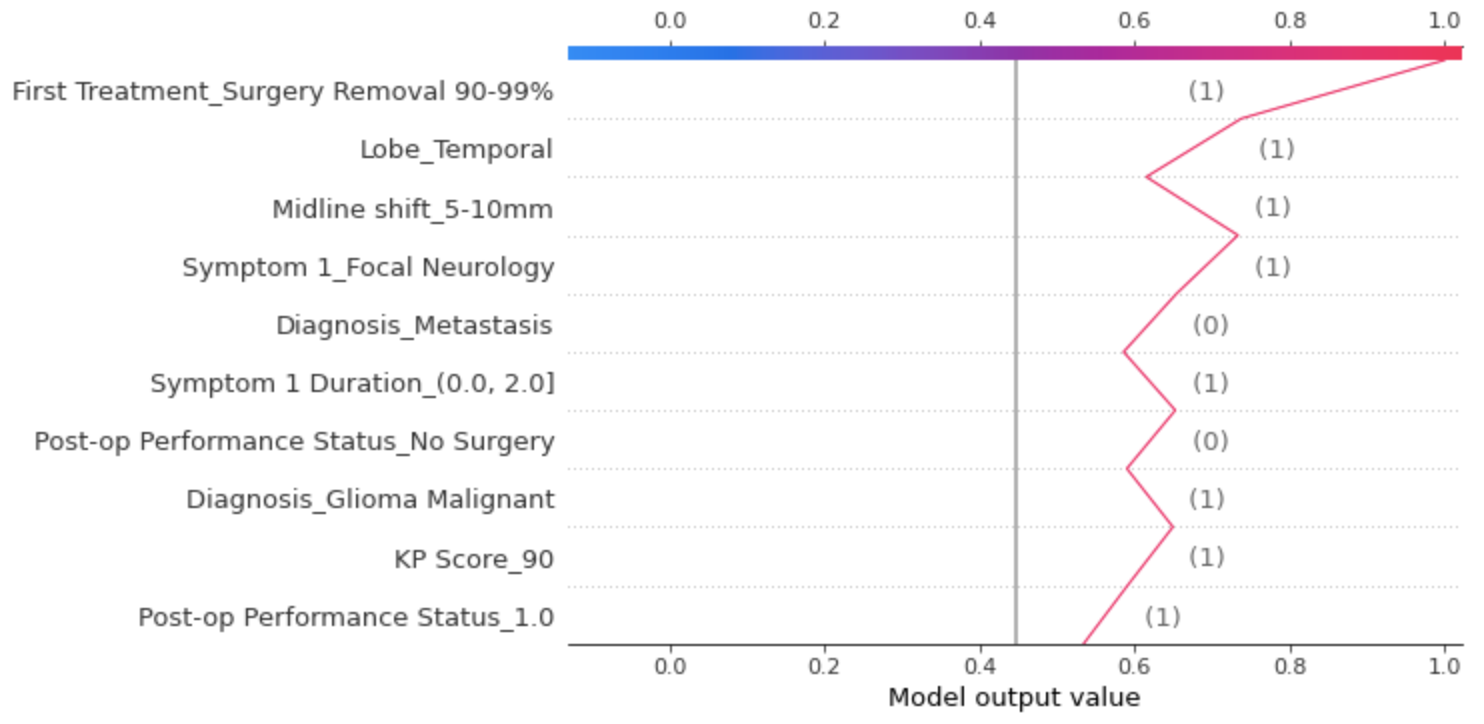}
    \caption{LR feature importance determined by SHAP.}
    \label{lr_shap}
\end{figure}

\begin{figure}[h!]
    \centering
    \includegraphics[width=1\linewidth]{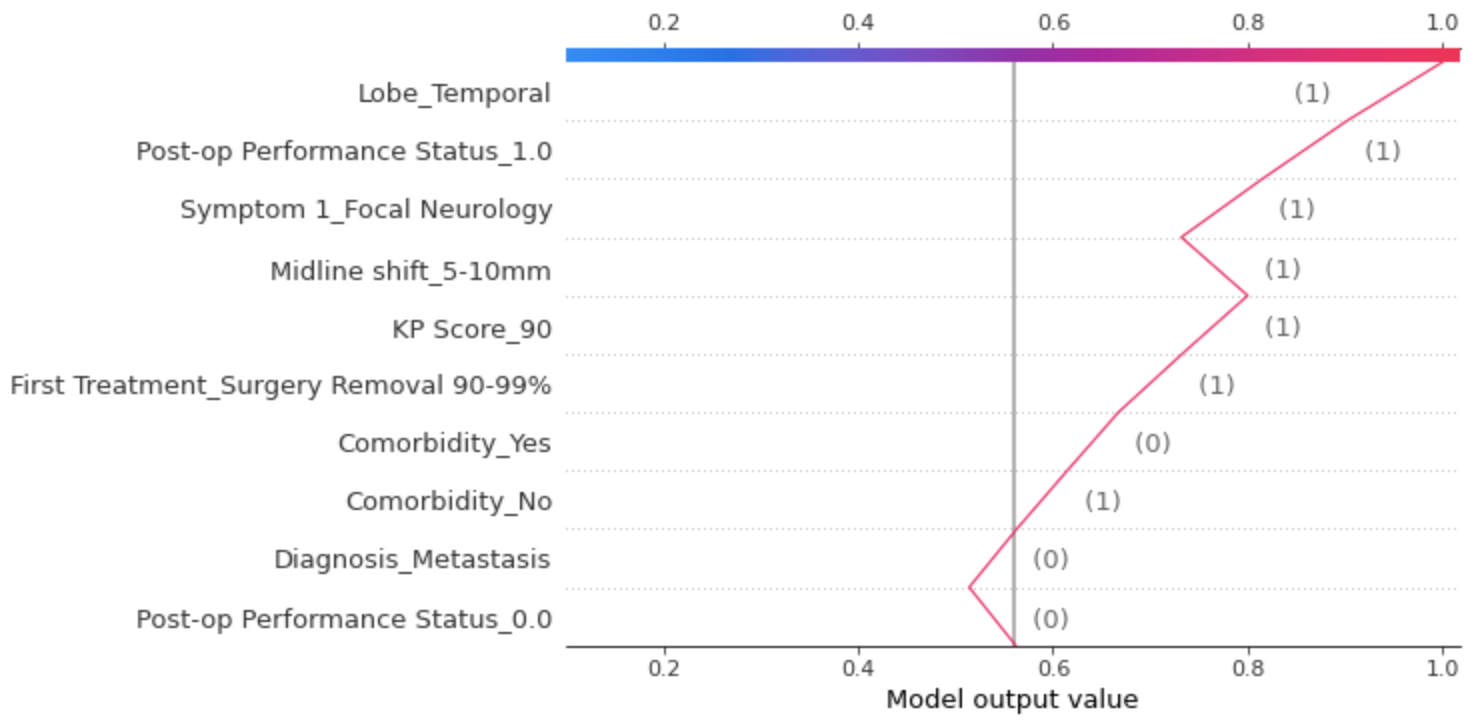}
    \caption{SVM feature importance determined by SHAP.}
    \label{svm_shap}
\end{figure}

\clearpage

\section*{Appendix B}\label{app_b}

\begin{figure}[h]
    \centering
    \includegraphics[width=1\linewidth]{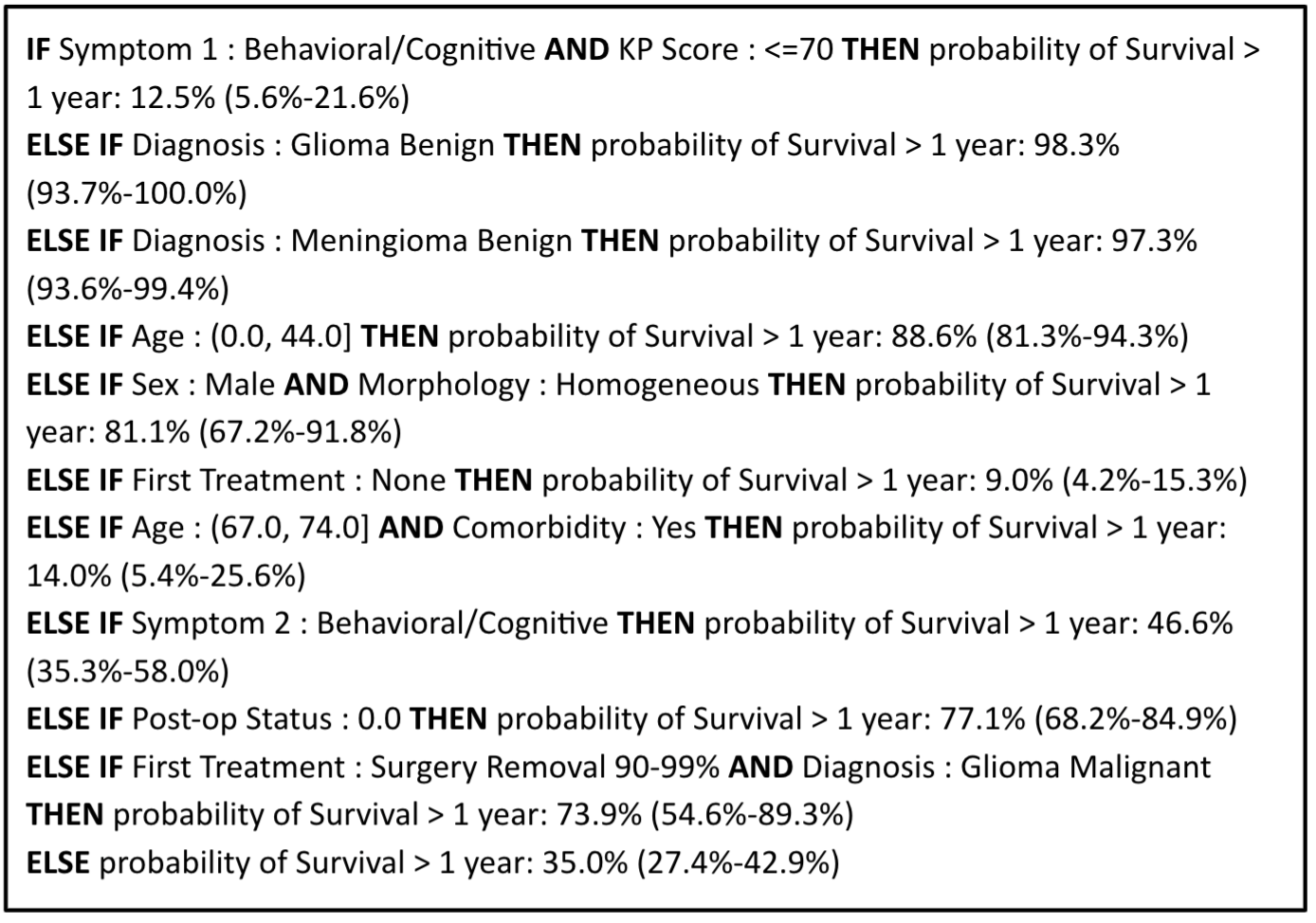}
    \caption{BRL-point estimate.}
    \label{brl2}
\end{figure}

\begin{figure}[h]
    \centering
    \includegraphics[width=1\linewidth]{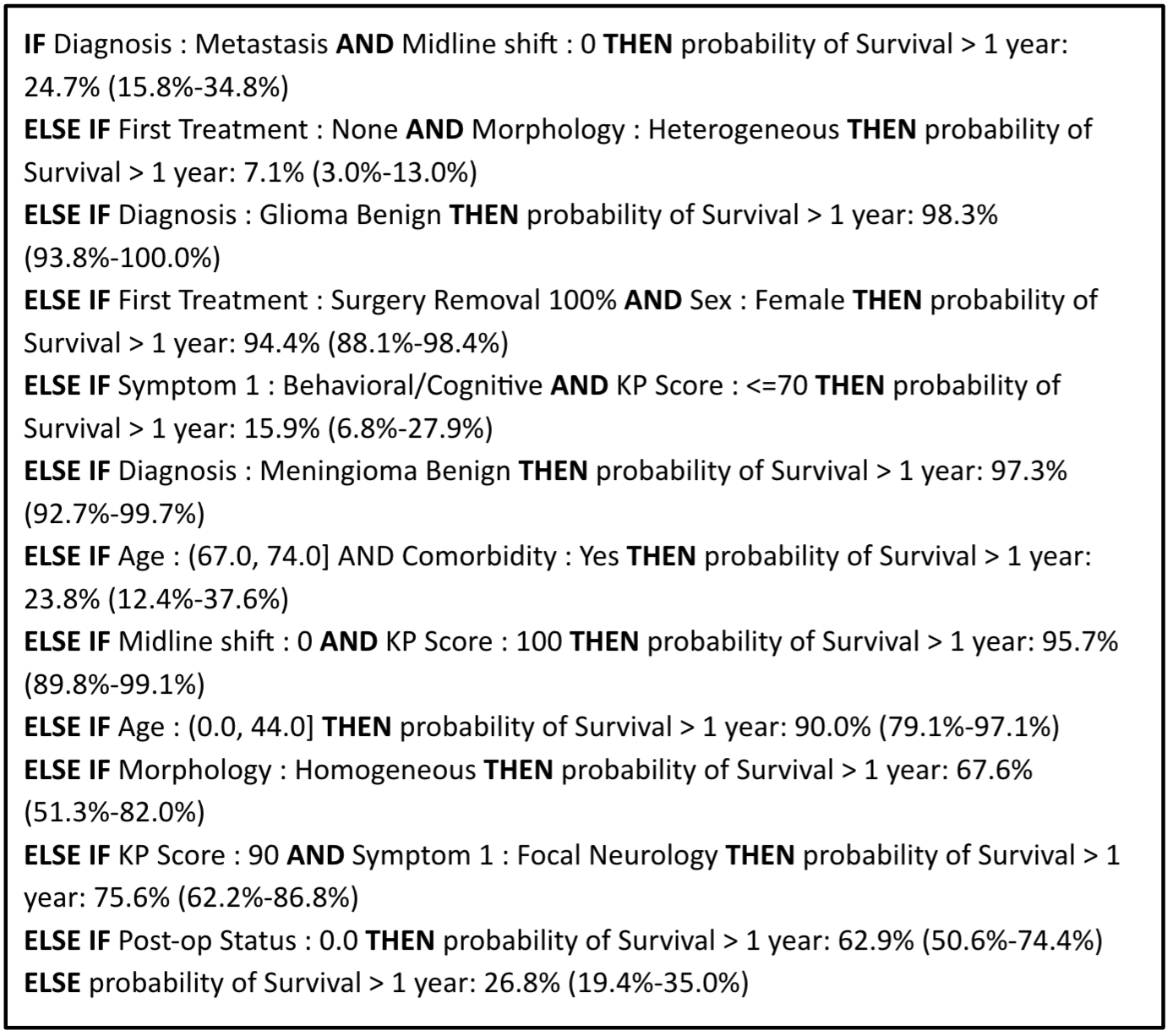}
    \caption{BRL-point estimate.}
    \label{brl3}
\end{figure}

\begin{figure}[h]
    \centering
    \includegraphics[width=1\linewidth]{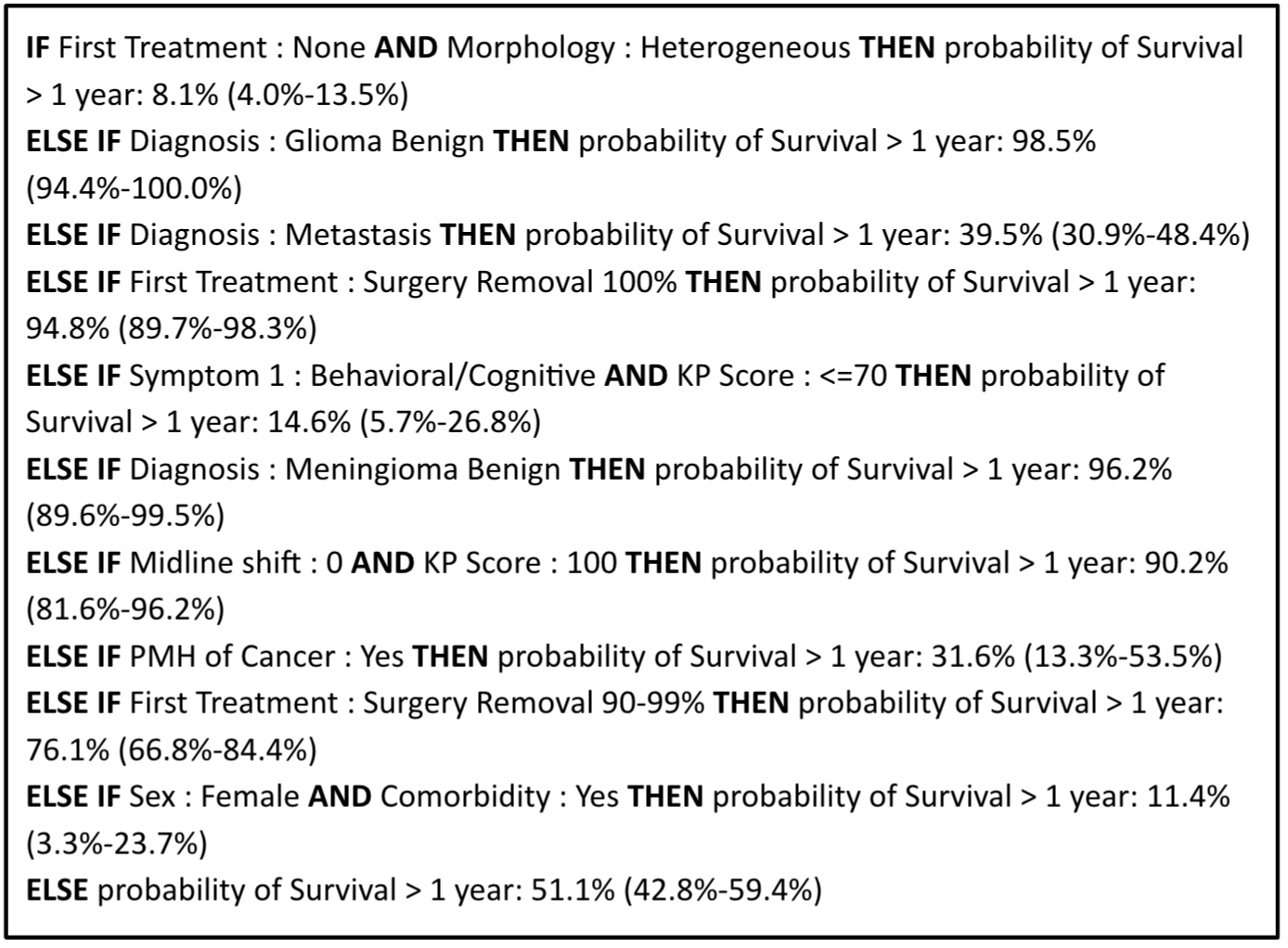}
    \caption{BRL-point estimate.}
    \label{brl4}
\end{figure}

\begin{figure}[h]
    \centering
    \includegraphics[width=1\linewidth]{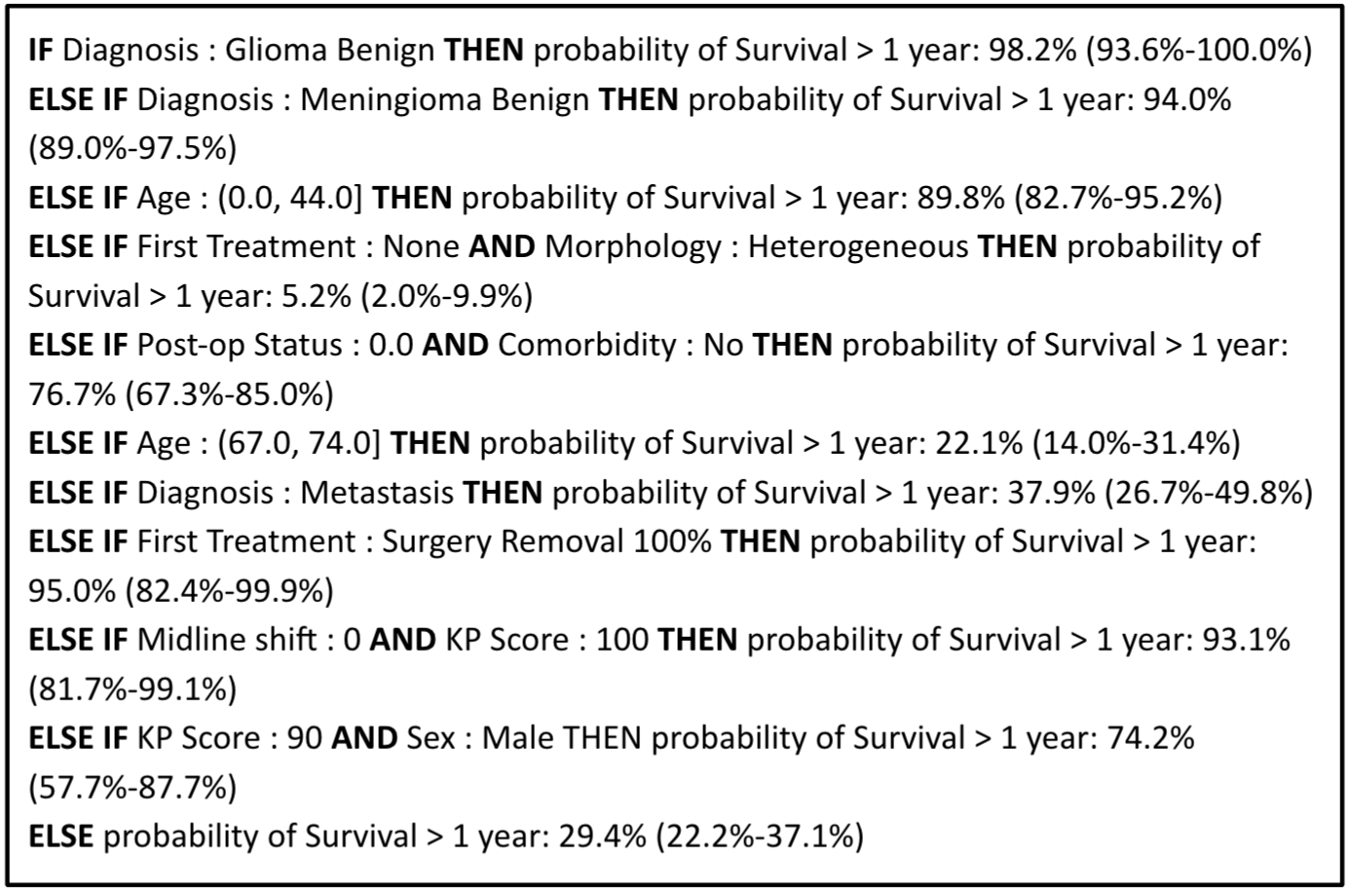}
    \caption{BRL-point estimate.}
    \label{brl5}
\end{figure}

\begin{figure}[h]
    \centering
    \includegraphics[width=1\linewidth]{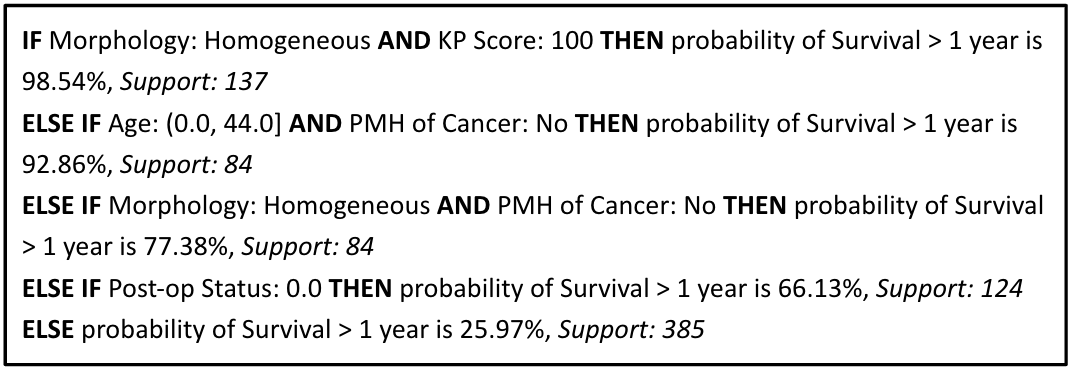}
    \caption{FRL-point estimate.}
    \label{frl2}
\end{figure}

\begin{figure}[h]
    \centering
    \includegraphics[width=1\linewidth]{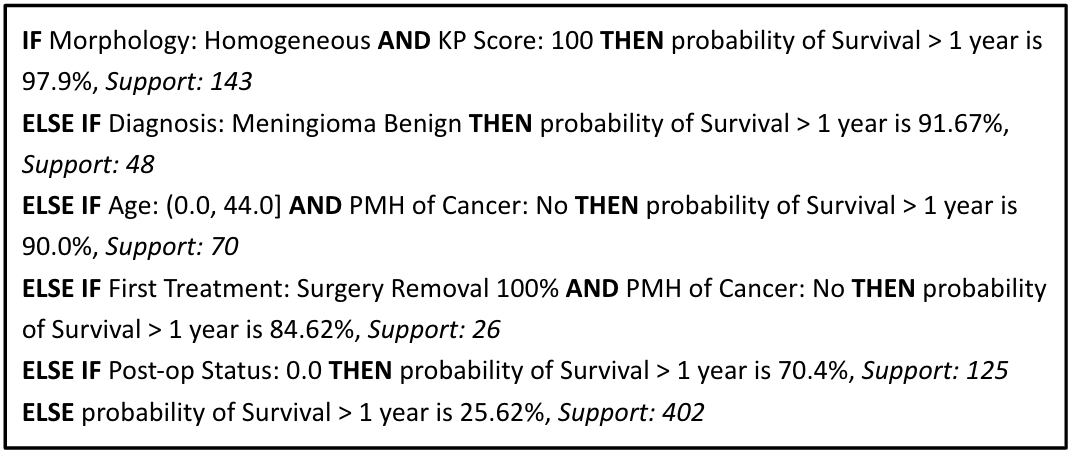}
    \caption{FRL-point estimate.}
    \label{frl3}
\end{figure}

\begin{figure}[h]
    \centering
    \includegraphics[width=1\linewidth]{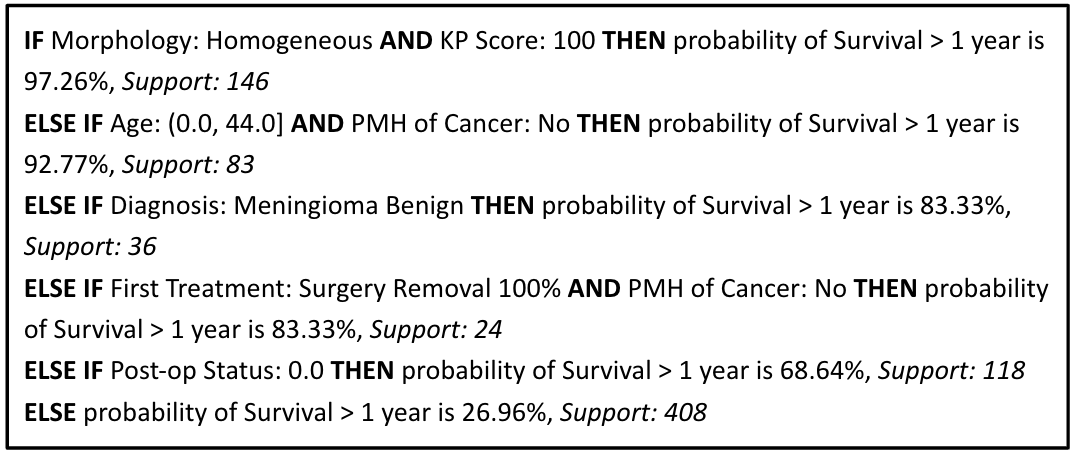}
    \caption{FRL-point estimate.}
    \label{frl4}
\end{figure}

\begin{figure}[h]
    \centering
    \includegraphics[width=1\linewidth]{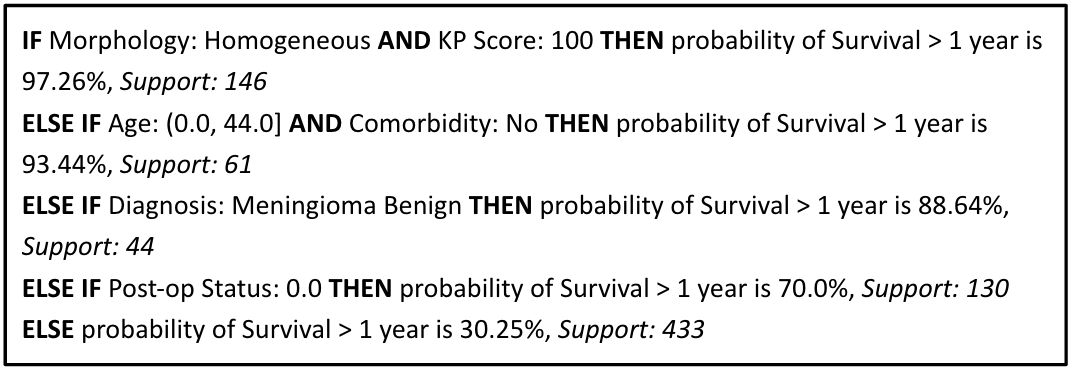}
    \caption{FRL-point estimate.}
    \label{frl5}
\end{figure}

\clearpage


\bibliography{mybibfile}

\end{document}